\documentclass[journal,12pt,onecolumn,draftclsnofoot,]{IEEEtran}
\IEEEoverridecommandlockouts
\usepackage{cite}
\usepackage{amsmath,amssymb,amsfonts}
\usepackage{graphicx}
\usepackage{textcomp}
\usepackage{xcolor}
\usepackage{multirow}
\usepackage{algorithm}
\usepackage{algorithmic}
\usepackage{color}
\usepackage{threeparttable}
\usepackage{booktabs} 
\usepackage{hyperref}
\usepackage{graphicx}
\usepackage{subfigure}
\usepackage{mathtools}
\usepackage{xcolor}% 颜色支持
\usepackage{colortbl}% 表格颜色支持
\usepackage{bm}
\usepackage{booktabs}
\usepackage{makecell}

\DeclarePairedDelimiterX\set[1]\lbrace\rbrace{#1}

\newcommand{\traj}{\mathcal{T}}
\usepackage{ulem} %to strike the words
\usepackage[binary-units]{siunitx}
\usepackage{cleveref}
\crefname{table}{Table}{Tables}
\crefname{figure}{Fig.}{Figs.}
\crefname{section}{Section}{Sections}

\usepackage{ifthen}
\newboolean{review}
\setboolean{review}{true} 

\ifthenelse{\boolean{review}}{
\usepackage{ulem}

\newcommand{\revdel}[1]{{\color{red}#1}}

\newcommand{\fixS}[1]{{\color{blue}\sout{#1}}}

}{
\newcommand{\revdel}[1]{{}}

\newcommand{\fixS}[1]{}

}
\newcommand*{\dif}{\mathop{}\!\mathrm{d}}

\usepackage[acronym]{glossaries}
\newacronym{ar}{AR}{Augmented Reality}
\newacronym{vr}{VR}{Virtual Reality}
\newacronym{mr}{MR}{Mixed Reality}
\newacronym{mse}{MSE}{Mean Squared Error}
\newacronym{drl}{DRL}{Deep Reinforcement Learning}
\newacronym{kc-td3}{KC-TD3}{Knowledge-assisted Constrained Twin-Delayed Deep Deterministic}
\newacronym{hci}{HCI}{Human-Computer Interaction}
\newacronym{cvar}{CVaR}{Conditional Value-at-Risk}
\newacronym{mtp}{MTP}{Motion-To-Photon}
\newacronym{embb}{eMBB}{Enhanced Mobile Broadband}
\newacronym{urllc}{URLLC}{Ultra-Reliable Low-Latency Communication}
\newacronym{mmtc}{mMTC}{Massive Machine Type Communication}
\newacronym{e2e}{E2E}{End-to-End}
\newacronym{aoi}{AoI}{Age of Information}
\newacronym{iiot}{IIoT}{Industrial Internet of Things}
\newacronym{mi}{MI}{Mutual Information}
\newacronym{lstm}{LSTM}{Long Short-Term Memory}
\newacronym{mlp}{MLP}{Multi-Layer Perception}
\newacronym{hdmi}{HDMI}{High-Definition Multimedia Interface}
\newacronym{apdo}{APDO}{accelerated primal-dual policy optimization}
\newacronym{cmdp}{CMDP}{Constrained Markov Decision Processes}
\newacronym{lstf}{LSTF}{Long Sequence Time-series Forecast}
\newacronym{vpg}{VPG}{Vanilla Policy Gradient}
\newacronym{trpo}{TRPO}{Trust Region Policy Optimization}
\newacronym{ddpg}{DDPG}{Deep Deterministic Policy Gradient}
\newacronym{ppo}{PPO}{Proximal Policy Optimization}
\newacronym{rmp}{RMP}{Riemannian Motion Policy}
\newacronym{crpo}{CRPO}{Constraint-Rectified Policy Optimization}
\newacronym{var}{VaR}{Value-at-Risk}
\newacronym{kpi}{KPI}{Key Performance Indicator}
\newacronym{dt}{DTs}{Digital twins}
\newacronym{agv}{AGV}{Autonomous Guided Vehicle}
\newacronym{isac}{ISAC}{Integral sensing and communication}
 \newacronym{pd}{PD}{proportional–derivative} 
 \newacronym{cppo}{C-PPO}{Constraint Proximal Policy Optimization} 
 \newacronym{gae}{GAE}{ Generalized Advantage Estimate}
\newacronym{ccdf}{CCDF}{Complementary Cumulative Distribution Function}
\newacronym{mam}{MA}{Moving Average Method}
\newacronym{dof}{DoF}{degrees of freedom}
\newacronym{ofdm}{OFDM}{Orthogonal Frequency-Division Multiplexing}
\newacronym{iot}{IoT}{Internet of Things}
\newacronym{dh}{D-H}{Denavit–Hartenberg}
\usepackage{mathrsfs}
\def\BibTeX{{\rm B\kern-.05em{\sc i\kern-.025em b}\kern-.08em
    T\kern-.1667em\lower.7ex\hbox{E}\kern-.125emX}}

\vspace{-0.6in}  

\begin{document}
\title{Task-Oriented Cross-System Design for Timely and Accurate Modeling in the Metaverse}

 \author{Zhen~Meng, \textit{Student Member}, \textit{IEEE}, Kan~Chen, \textit{Student Member}, \textit{IEEE}, Yufeng~Diao, \textit{Student Member}, \textit{IEEE}, Changyang~She, \textit{Senior Member}, \textit{IEEE}, Guodong~Zhao, \textit{Senior Member}, \textit{IEEE}, Muhammad~Ali~Imran, \textit{Fellow}, \textit{IEEE}, and Branka~Vucetic, \textit{Life Fellow}, \textit{IEEE}

\thanks{Z. Meng, K. Chen, Y. Diao, G. Zhao, and M. A. Imran are with the James Watt School of Engineering, University of Glasgow, UK. (e-mail: z.meng.1@research.gla.ac.uk; k.chen.1@research.gla.ac.uk; y.diao.1@research.gla.ac.uk; guodong.zhao@glasgow.ac.uk; muhammad.imran@glasgow.ac.uk)}
\thanks{C. She and B. Vucetic are with the School of Electrical and Information Engineering, University of Sydney, Australia. (e-mail: shechangyang@gmail.com; branka.vucetic@sydney.edu.au)}
\thanks{Corresponding author: Changyang She.}
}

\maketitle 
\pagestyle{empty} 
\thispagestyle{empty}

\begin{abstract}
In this paper, we establish a task-oriented cross-system design framework to minimize the required packet rate for timely and accurate modeling of a real-world robotic arm in the Metaverse, where sensing, communication, prediction, control, and rendering are considered. To optimize a scheduling policy and prediction horizons, we design a \gls{cppo} algorithm by integrating domain knowledge from relevant systems into the advanced reinforcement learning algorithm, \gls{ppo}. Specifically, the Jacobian matrix for analyzing the motion of the robotic arm is included in the state of the \gls{cppo} algorithm, and the \gls{cvar} of the state-value function characterizing the long-term modeling error is adopted in the constraint. Besides, the policy is represented by a two-branch neural network determining the scheduling policy and the prediction horizons, respectively. To evaluate 
our algorithm, we build a prototype including a real-world robotic arm and its digital model in the Metaverse. The experimental results indicate that domain knowledge helps to reduce the convergence time and the required packet rate by up to $50$\%, and the cross-system design framework outperforms a baseline framework in terms of the required packet rate and the tail distribution of the modeling error.

\end{abstract}
  
\begin{IEEEkeywords}
Task-oriented cross-system design, scheduling, prediction, constraint deep reinforcement learning, Metaverse.
\end{IEEEkeywords}

\glsresetall

% Metaverse = convergence of physical, augmented and virtual reality

\section{Introduction}
As a digital world that mirrors the physical world and generates feedback for human users in real-time, the Metaverse can blur the lines between the physical and digital worlds and revolutionize how humans communicate and interact with each other~\cite{lee2021all}. To achieve this goal, the timely and accurate modeling of real-world devices/humans in the Metaverse is critical for user experience. The communication and computing latency and digital model distortion will lead to chaotic interactions and user dizziness~\cite{583063}. For some mission-critical applications assisted by the Metaverse, such as remote healthcare and factory automation, slight out-of-synchronization between a real-world device and its digital model may cause serious consequences~\cite{laaki2019prototyping}. 

Efficiently replicating the real world in the Metaverse brings significant challenges in communications. While 5G networks have significantly improved latency, reliability, data rate, and connection density, it still falls short of satisfying the demands of the Metaverse~\cite{3GPP}. One of the examples is that 5G New Radio can achieve $1$~ms latency and $10^{-5}$ packet error probability in the radio access networks (RANs), but does not guarantee end-to-end delay and reliability. Jitter is another issue that can lead to inaccurate modeling, which may be caused by 5G RANs and 5G core networks. Furthermore, there is a mismatch between communication \glspl{kpi}, i.e., latency, reliability, jitter, and throughput, and the \gls{kpi} requirements of diverse tasks in the Metaverse, such as modeling error, haptic feedback distortion, and semantic segmentation errors, which will lead to poor user experience and low resource utilization efficiency.

On the other hand, the performance of modeling in the Metaverse is not solely determined by communication networks. Other systems, including sensing, prediction, control, and rendering, can also have significant impacts on \gls{e2e} latency and accuracy. Designing these systems separately results in strictly sub-optimal solutions and may fail to meet the task-oriented KPI requirements. Thus, cross-system design has been investigated in the recent literature, such as prediction and communication co-design~\cite{girgis2021predictive, hou2019prediction} and sampling and communication co-design~\cite{15450, 8812616}. They have shown significant gains of the cross-system design, but they have also revealed potential issues. For example, cross-system models could be analytically intractable, and the complexity of cross-system problems can be much higher than problems in separate design approaches. Thus, novel methodologies are needed for cross-system design.

The first step toward cross-system design is to formulate a problem that takes the relevant systems into account. Nevertheless, it could be difficult to obtain closed-form expressions of the objective function and the constraints as some of the \glspl{kpi} are analytically intractable. Although we can use some approximations to formulate the problem, it is generally non-convex or NP-hard. This motivates us to develop data-driven deep learning approaches, where the policy to be optimized is represented by a neural network. \gls{drl} is a promising method for training the neural network. For example, \gls{ppo} is developed to optimize policies with a discrete action space~\cite{schulman2017proximal}. More recently, the primal-dual method and \gls{crpo} were introduced into \gls{drl} for solving constrained problems \cite{liang2018accelerated, 01629 ,xu2021crpo}. It is worth noting that a straightforward implementation of \gls{drl} algorithms may not work \cite{gu2021knowledge}. Integrating domain knowledge from relevant systems into \gls{drl} algorithms is essential for the success of \gls{drl}~\cite{she2021tutorial} in practical applications.

\subsection{Related Work}

\subsubsection{Gaps between 5G/6G and Metaverse}
The concept of the Metaverse was initially introduced in Neil's book, \textit{Snow Crash}~\cite{stephenson2003snow}, coinciding with the development of virtual physical fusion technology. Notable pioneering contributions have been made in various applications of the Metaverse, including the game networking~\cite{01672}, autonomous vehicles~\cite{06838}, \gls{iot} devices~\cite{9865226}, and education~\cite{01512}. The above work has yielded valuable insights that could be leveraged to advance the development of the Metaverse, particularly with regard to the conception and refinement of forthcoming communication systems and network architectures. However, there still exists a research gap between 5G/6G and the Metaverse: 1) The KPIs in 5G such as latency, reliability, and throughput cannot fulfill the requirements of diverse tasks in the Metaverse~\cite{lee2021all}. This misalignment leads to significant challenges in future communication system design for the Metaverse. For example, there is no closed-form relation between the KPIs in 5G and the requirement of tasks in the Metaverse. As a result, it is difficult to guarantee End-to-End (E2E) performance requirement~\cite{9681718}. 2) Future 6G networks should support a large number of devices with diverse KPIs in the Metaverse by leveraging distributed computing, storage, and communication resources in local devices, edge servers, and central servers~\cite{multi-tier}. A hierarchical communication network architecture with a strong cloud server and multiple edge servers is promising, where edge servers generate quick responses to mobile devices and update critical messages to the cloud. 3) Existing HCI, sensing, communication, and computing systems are developed separately~\cite{khan2023metaverse}, which leads to sub-optimal solutions, brings extra communication overhead for coordinating multiple tasks and cannot meet task-oriented KPIs~\cite{8070468}.
\subsubsection {Timely,  Efficiently and Accurate Modeling for Digital Twin and Metaverse}
In 2012, NASA clarified the concept of Digital Twins and defined it as ``integrated multiphysical, multiscale, probabilistic simulations of an as-built vehicle or system using the best available physical models, sensor updates, and historical data”~\cite{glaessgen2012digital}. However, the communication, computing, and storage resources in the existing IT and networking infrastructures are insufficient to support of diverse applications associated with digital twins and the Metaverse. Some existing studies focused on the timely,  efficient, and accurate modeling of digital twins in the Metaverse. \cite{13445,14686,03300,9491087, hashash2022edge, 9865226}. In~\cite{9491087}, a deep reinforcement learning approach was developed to solve the placement and migration problems of digital twins to minimize the synchronization delay with the help of edge computing. The authors of~\cite{9865226} proposed a resource allocation algorithm for synchronizing Internet of Things devices with their digital models in the Metaverse by using a game-theoretic framework. In~\cite{hashash2022edge}, the authors proposed an edge continual learning framework that can accurately synchronize a physical object with evolving affinity with its digital twin. Instead of using a centralized framework, a distributed Metaverse framework was developed in~\cite{9865226} for synchronizing multiple digital twins, known as sub-Metaverses. The above work revealed some fundamental tradeoffs between resource utilization efficiency and \glspl{kpi} in the Metaverse. Nevertheless, coordinating communications, computing, and storage sources for diverse Metaverse applications remains an open problem due to its high complexity, especially when the scale of the system is large.

\subsubsection{Task-Oriented Cross-System Design For Metaverse} While 5G can support enhanced broadband services with high data rates, ultra-reliable and low-latency communications, and massive machine-type communications~\cite{3GPP}, it still falls short of satisfying task-oriented KPIs in the Metaverse, such as modeling error, haptic feedback distortion, and semantic segmentation error~\cite{yu20236g}. To fill the gap between the communication \glspl{kpi} and the \gls{kpi} requirements of specific tasks, task-oriented cross-system design is a promising approach. The authors of~\cite{de2023goal} considered an Age-of-Loop metric for the remote control of autonomous guided vehicles, and proposed a goal-oriented wireless solution that adjusts the data rate to achieve high control accuracy. Their results showed that with the goal-oriented \gls{kpi}, it is possible to achieve higher accuracy than the commonly used communication \glspl{kpi}, such as Age-of-Information. 
In~\cite{shao2021learning}, the authors proposed a learning-based communication scheme that optimizes feature extraction, source coding, and channel coding in a task-oriented manner to achieve low-latency inference for image classification. The experimental results of this work indicated that task-oriented communication achieves a better rate-distortion tradeoff than baseline methods. More recently, the authors of \cite{00969} developed \gls{e2e} task-oriented resource management by integrating sensing, computing, and communication processes into a joint design framework, where the artificial intelligence model is split and executed on edge servers for low-latency intelligent services. To improve the user experience in immersive Metaverse applications, the authors of~\cite{01471} proposed a user-centered joint optimization approach to optimize frame generation location, transmission power, and channel access arrangement. These studies indicated that by task-oriented cross-system design, it is possible to provide a better user experience and achieve higher resource utilization efficiency. However, Metaverse is expected to handle large amounts of heterogeneous data from various sources and formats. How to develop task-oriented data representation and ontology models that enable efficient data exchange, integration, and interpretation within the Metaverse context still needs further investigation~\cite{jagatheesaperumal2023semantic}. Furthermore, the cross-system problems are non-convex or NP-hard in general. Finding a near-optimal resource management solution with low complexity remains a challenging issue.

\subsection{Contributions}
In this paper, we aim to address the following fundamental issues: 1) How to eliminate the gap between traditional communication \glspl{kpi} defined in the 5G standard and user-centered task-oriented \glspl{kpi} in the Metaverse? 2) How to implement cross-system design in the Metaverse, including sampling, communication, prediction, control, and rendering? 3) How to utilize cross-system domain knowledge to guide the \gls{e2e} training of \gls{drl} algorithms? The main contributions of this paper are summarized as follows:

\begin{itemize}
\item We establish a task-oriented cross-system design framework, where sensing, communication, prediction, control, and rendering are jointly considered for modeling a robotic arm in the Metaverse. The scheduling policy and the prediction horizon are jointly optimized to minimize the required packet rate to guarantee a modeling error constraint.

\item We propose a \gls{cppo} algorithm by integrating domain knowledge into the advanced \gls{ppo} algorithm, and further train the policy using the \gls{cppo} algorithm. Specifically, 1) the Jacobian matrix, which is widely used for analyzing the motion of robotic arms, is included in the state of \gls{drl} to improve the training efficiency of \gls{cppo}. 2) The \gls{cvar} of the state-value function that characterizes the long-term modeling error is applied to formulate the constraint. 3) A two-branch neural network is developed to determine the scheduling policy and the prediction horizons.

\item We build a prototype system including a real-world robotic arm and its digital model in the Metaverse, where the Nvidia Issac Gym platform is used. Extensive experiments are carried out in the prototype to evaluate the proposed task-oriented cross-system design approach. The experimental results show that our ~\gls{cppo} outperforms several benchmark algorithms in terms of convergence time, stability, packet rate, and modeling error, and the cross-system design framework outperforms a baseline framework in terms of the required packet rate and the tail distribution of tracking errors.
\end{itemize}

The rest of this paper is organized as follows.
In \cref{sec:method}, we propose the task-oriented cross-system design framework where all subsystems, i.e., sensing, communication, 
 reconstruction, prediction, control, and rendering, are elaborated in detail.
In \cref{sec:kctd3}, we develop the \gls{cppo} algorithm to optimize the scheduling and prediction policy while minimizing the communication load under the constraint of \gls{cvar}. 
\cref{sec:results} describes the prototype and provides performance evaluations.
Finally, \cref{sec:conclusions} concludes this paper.
 
\section{Task-Oriented Cross-System Design}
\label{sec:method}

In this section, we propose a task-oriented cross-system design framework for timely and accurate modeling in the Metaverse. The specific goal is to build a digital model of a real-world robotic arm for real-time monitoring and control. 

\begin{table}[t] % Guodong: table should always be the top of the page.
  \begin{center}
    \caption{SUMMARY OF MAIN SYMBOLS}
    \resizebox{\textwidth}{!}{
    \begin{tabular}{|c|l|c|l|} % <-- Alignments: 1st column left, 2nd middle and 3rd right, with vertical lines in between
      \hline
      \rowcolor[RGB]{200,200,200} 
      \textbf{Symbol} & \textbf{Explanation}
      &\textbf{Symbol} & \textbf{Explanation}\\
      \hline
      \hline
      $t_s$ & Duration of time slot 
      & ${\mathcal{P}}(t)$ & Pose of real-world robotic arm in the $t$-th time slot\\
      \hline
      \rowcolor[RGB]{230,230,230}
      $I$ & The number of joints 
      &$\bm{\eta}_c(t)$ & Unit vector of rotation axis\\
      \hline
      $\traj(t)$ & Trajectory of the real-world robotic arm in the $t$-th time slot 
      &$\psi_{\eta}(t) $ & Angle of rotation\\
      \hline
      \rowcolor[RGB]{230,230,230}
      $\tau_i(t)$ & Angle value of the $i$-th joint measured in the $t$-th time slot
      & $F_f(\cdot)$ & Forward kinematics of real-world robotic\\
      \hline
      $x_i(t)$ &  Whether the i-th joint is scheduled to transmit in the $t$-th time slot
      &$\check{P}(t)$ & Pose of virtual-world robotic arm in the $t$-th time slot\\
      \hline
      \rowcolor[RGB]{230,230,230}
      $X(t)$ & Decision of the scheduler in the $t$-th time slot
      & ${\bf{e}}(t)$ & Modeling error in the $t$-th time slot\\
      \hline
      $y_i(t)$ & Indicators of whether packet arrivals in the $t$-th time slot 
      &$\omega_1$, $\omega_2$ & Weighting coefficients of the modeling error\\
      \hline
      \rowcolor[RGB]{230,230,230}
      $\dot{\mathbb{S}}(t)$ & Set of angles received by the Metaverse 
      & $\pi_{\theta}$ & Scheduling and prediction policy \\
      \hline
      $W_l$ & Historical observation window for reconstruction
      & $\theta$ & Parameters of policy \\
      \hline
      \rowcolor[RGB]{230,230,230}
      $\bar{\traj}(t_0)$ & Reconstructed trajectory in the $t_0$ time slot
      & ${\bf{a}}_t$ & Action taken in the $t$-th time slot\\
      \hline
      $F_l(\cdot,\theta_l)$ &Function of reconstruction
      & ${\bf{s}}_t$ & State\\
      \hline
      \rowcolor[RGB]{230,230,230}
      $\theta_l$ & Parameters of reconstruction
      &$a{_{t,i}}^{[1]}, a{_{t,i}}^{[2]}$  & Action of the $i$-th joint taken in the $t$-th time slot\\
      \hline
      $W_p$ & Input length of the prediction 
      &$\rho{_{t,i}}^{[1]}, \rho{_{t,i}}^{[2]}$  & Distribution of action in the $t$-th time slot\\
      \hline
      \rowcolor[RGB]{230,230,230}
      $H$ & Output length of the prediction 
      & $\mathcal{L}_c$, $\mathcal{L}_r$ &Loss functions of policy network\\
      \hline
      $F_p(\cdot,\theta_p)$ & Function of prediction
      & $A(\bf{s},\bf{a})$ & Advantage function \\
      \hline
      \rowcolor[RGB]{230,230,230}
      $\theta_p$ & Parameters of the prediction function
      & $Q^{\pi_{\theta}}({\bf{s}},\bf{a})$ & State-action value under policy \\
      \hline
      $L_{\text{p}}$ & MSE loss of prediction
      & $V^{\pi_\theta}({\bf{s}})$ & State value function\\
      \hline
      \rowcolor[RGB]{230,230,230}
      $\hat{\traj}(t)$ & Prediction trajectory in the $t$-th time slot
      & $\mathcal{J}(\cdot)$ &Jacobian matrix\\
      \hline
      $\hat{\tau}_i(t)$ & Prediction trajectory of the $i$-th joint in the $t$-th time slot
      &$L_1$, $L_2$, $L_3$ & Length of joint links \\
      \hline
      \rowcolor[RGB]{230,230,230}
       $Z(t)$ & Optimal prediction horizon
       &  $\phi(t)$ & Angle between $x$-axis and $x'$-axis\\
      \hline
      $z_i(t)$ & Optimal prediction horizon of the $i$-th joint
      & $r({\bf{s}}_t,\bf{a}_t)$ & Instantaneous reward\\
      \hline
      \rowcolor[RGB]{230,230,230}
      $N_c$ &  Time interval of control command generation
      &$c({\bf{s}}_t,\bf{a}_t)$ & Instantaneous cost\\
      \hline
      $N_r$ & Processing time of each image
      &$R^{\pi_\theta}$ & Long-term reward\\
      \hline
      \rowcolor[RGB]{230,230,230}
      $k_p$ & Proportional parameter of PD controller
      &$C^{\pi_\theta}$ & Long-term cost\\
      \hline
      $k_d$ & Derivative parameter of PD controller
      & $\gamma$ & Discount factor\\
      \hline
      \rowcolor[RGB]{230,230,230}
      $\widetilde{\traj}(t)$ & Control results in the $t$-th time slot
      &$\Gamma_c$ & Constraint of modeling error \\
      \hline
       $\widetilde{\tau}_i(t)$ & Control result of the $i$-th joint in the $t$-th time slot
      & $\text{CVaR}_{\alpha}(\cdot)$&  CVaR function\\
      \hline
      \rowcolor[RGB]{230,230,230}
      $F_r(\cdot,\theta_r)$ & Function of rendering 
      & 1-$\alpha$ & Confidence level of CVaR\\
      \hline
      $\theta_r$ & Parameters of rendering function
      & $v$ & Multiplier of CVaR\\
      \hline
      \rowcolor[RGB]{230,230,230}
      $\check{\traj}(k)$ & The rendered trajectory
      & $\beta$ &  Learning rate of CVaR \\
      \hline
    \end{tabular}
    }
  \end{center}
\end{table}

\subsection{Framework}
\begin{figure}
\centering
\includegraphics[scale=0.35]{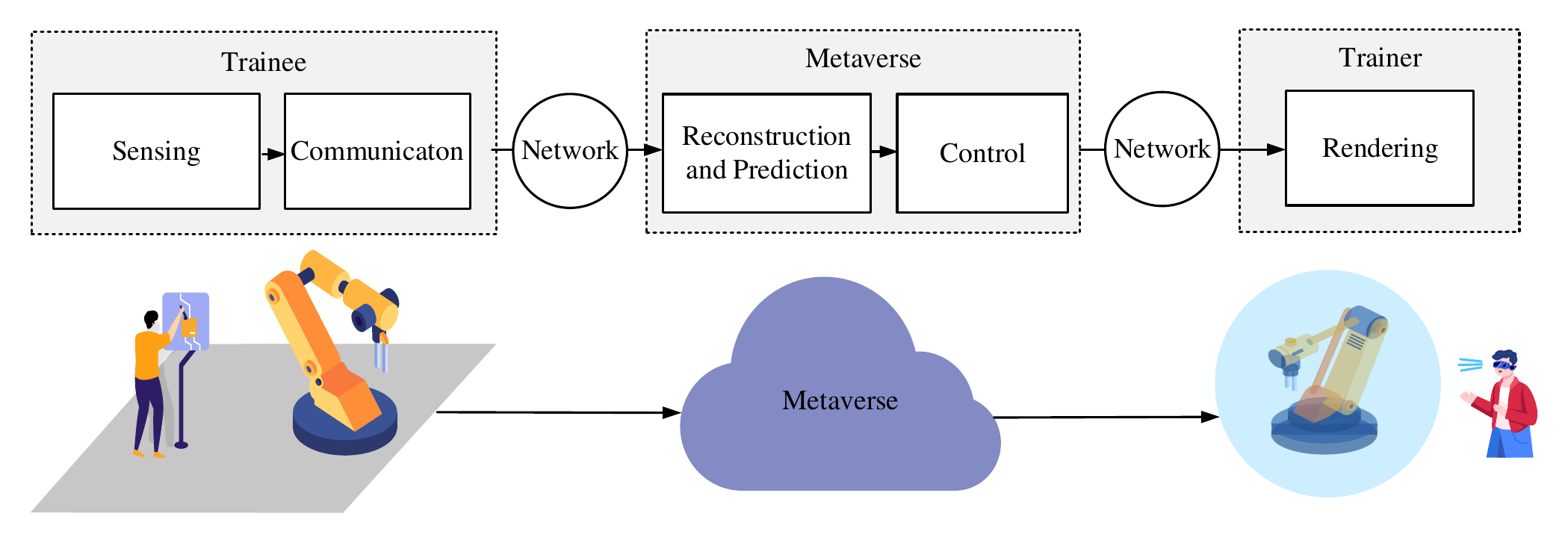}
\caption{Proposed task-oriented cross-system design framework for modeling a robotic arm in the Metaverse, where sensing, communication, reconstruction, predication, control and rendering are considered.}
\label{Illustration of system model}
\end{figure}
The framework is shown in Fig.~\ref{Illustration of system model}, where the real-world robotic arm with multiple joints is controlled by a user (a student or a trainee) for some tasks for example training in healthcare. The trajectory (the angles of all the joints) of the real-world robotic arm is measured by the built-in sensors. Then, the trajectory is sampled and transmitted to the Metaverse in a cloud server, where the sampled data are used to reconstruct the historical trajectory and predict the future trajectory. Here, the digital model in the Metaverse is controlled by the predicted trajectory rather than the reconstructed trajectory to compensate for delays caused by different components across systems. Finally, the digital model in the Metaverse is rendered and presented to another user (an expert or a trainer) via a computer screen or VR/AR headset. It is worth noting that each joint has its own state, and the states of the joints are interdependent. They need to collaborate with each other to accomplish the task. In addition, the total communication resources shared by all the joints are limited. Thus, it is possible to extend our system into multi-sensor scenarios. 

The \gls{e2e} \gls{mtp} latency is defined by the delay between the movement of the real-world robotic arm and the movement of its digital model in the Metaverse. Thus, it includes communication delay, computation delay, control delay, and rendering delay. By optimizing the prediction horizon and the scheduling policy\footnote{The scheduling policy determines which joints will be scheduled for data transmission.}, we minimize the communication overhead subject to constraints on the modeling accuracy and the \gls{mtp} latency.

\begin{figure}
\centering
\includegraphics[scale=0.5]{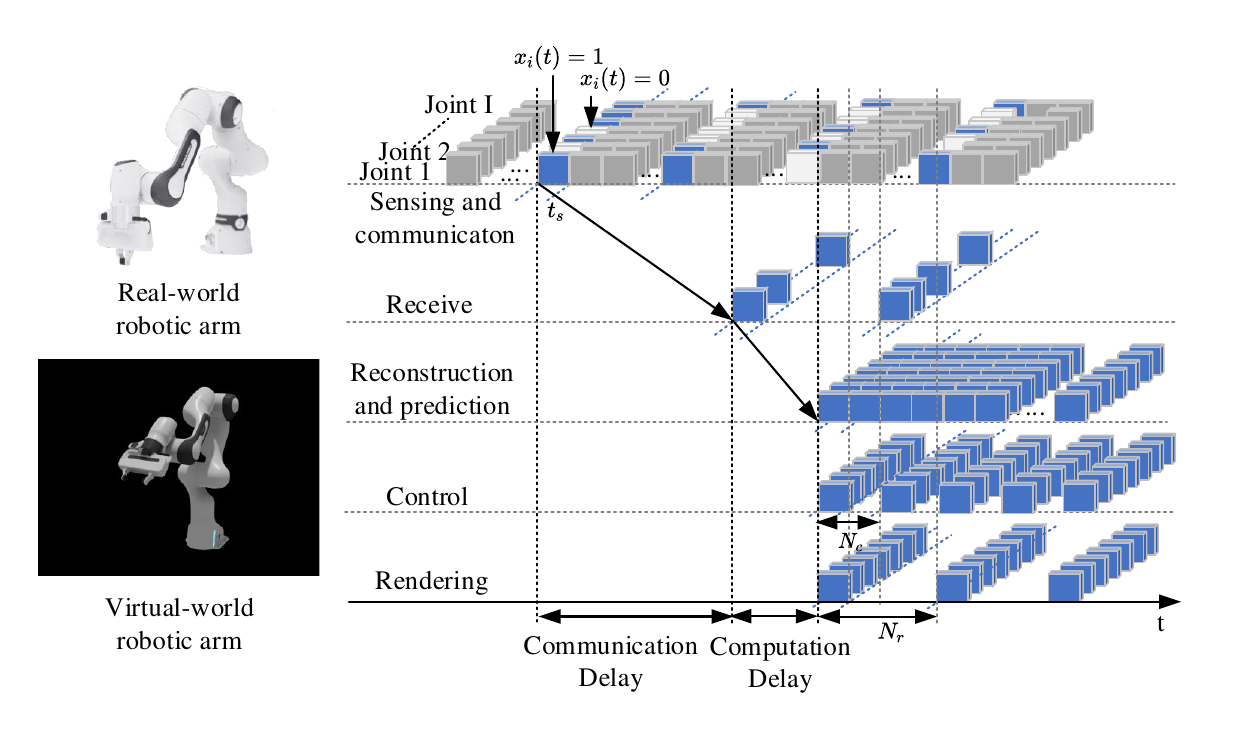}
\caption{The timing sequence of the proposed framework, where the modeling accuracy and the MTP latency need to be satisfied.}
\label{time flow}
\end{figure}

Fig. \ref{time flow} illustrate the timing sequence of the proposed framework. The data is generated from the built-in sensors at the physical robotic arm. Then, the communication module conducts scheduling and sends the selected data to a computer server via a network. The server conducts data reconstruction and prediction, and then controls the digital model of robotic arm. Finally, the digital model was rendered\footnote{To simplify the system, we assume that the rendering takes place at the server and the human user (trainer) directly interacts with the digital model via human-computer interface. } and presented to a human (trainer) via a VR headset (or a screen). In the following, we will introduce each component:

\subsubsection{Sensing and Communications} Time is discretized into slots with a duration of $t_{s}$. The built-in sensors measure $I$ joint angles in each time slot. Let $\traj(t) = [\tau_1(t), ...,\tau_I(t)]$ be the trajectory of the real-world robotic arm, where $\tau_i(t)$, $i=1,2,3,...,I$, is the angle value of the $i$-th joint measured in the $t$-th time slot. We consider a scheduling policy in the communication system, where the indicator, $x_i(t)$, represents whether the $i$-th joint is scheduled for data transmission in the $t$-th time slot, $i = 1,...,I$. If the $i$-th joint is not scheduled, then $x_i(t) = 0$. Otherwise, $x_i(t) = 1$, and one packet will be transmitted to the Metaverse. The decision of the scheduler in the $t$-th time slot is denoted by $X(t) = [x_1(t), x_2(t),...,x_I(t)]$. The total number of packets to be transmitted in the communication systems in the $t$-th time slot is given by $\sum_{i=1}^{I}x_i(t)$.

\subsubsection{Reconstruction} 
To reconstruct the trajectory from sampled data, we use the linear interpolation and extrapolation method, which is widely used in the existing literature and can be easily implemented in our system~\cite{scaglia2010linear}. The indicators of packet arrivals in the $t$-th time slot at the Metaverse are denoted by $y_i(t)$, $i=1,2,..., I$. If a packet from the $i$-th joint arrived at the Metaverse in the $t$-th time slot, then $y_i(t) = 1$. Otherwise, $y_i(t) = 0$. From the arrived packets, the set of joint angles obtained by the Metaverse in the $t$-th time slot is denoted by $\mathbb{S}(t) ={\{\tau_i(t) \mid y_i(t) = 1,\ i=1,...,I\}}$. 
In a certain time slot $t_0$, the cloud server reconstructs the trajectory of the robotic arm from the received joint angles in a historical observation window with a duration of $W_l$ time slots. The reconstruction algorithm is given by  
\begin{align}\label{reconstruct function}
\bar{\traj}(t_0) = F_l(\mathbb{S}(t), \theta_l \mid t \in [t_0-W_l, t_0-W_l+1,...,t_0-1]),
\end{align}
where $\bar{\traj}(t_0) \in \mathbb{R}^{1\times I}$ is the reconstructed trajectory, $F_l(\cdot,\theta_l)$ is the reconstruction function, and $\theta_l$ is the interpolation and extrapolation parameters.

\subsubsection{Prediction}
To compensate for the \gls{mtp} latency, we propose to use the attention-mechanism-based predictor, referred to as \textit{Informer}, to predict the future trajectory from the historical trajectory \cite{zhou2021informer}. %for its ultra-low computation overhead and extraordinary prediction accuracy on \gls{lstf}. 
The lengths of the input and output trajectories are denoted by $W_p$ and $H$. The values of $W_p$ and $H$ are determined by the auto-correlation coefficient of the trajectories \cite{ross1995stochastic}. We denote the prediction results for trajectory in the $t$-th time slot by $\hat{\traj}(t)=[\hat{\tau}_1(t), \hat{\tau}_2(t),...,\hat{\tau}_I(t)]$. In a certain time slot $t_1$,  the prediction algorithm can be expressed as follows,
\begin{align}\label{prediction}
[\hat{\traj}(t_1+1), \hat{\traj}(t_1+2), ...,\hat{\traj}(t_1+H)] =F_p([\bar{\traj}(t_1-W_p), \bar{\traj}(t_1-W_p+1), ...,\bar{\traj}(t_1)], \theta_p),
\end{align}
 where $F_p(\cdot,\theta_p)$ is the prediction function and $\theta_p$ represents the parameters of this function. The loss function of the prediction algorithm is \gls{mse} between the output trajectory and the ground truth, which is given by
% \begin{align}\label{prediction}
% {L_{\rm p}} = \frac{1}{H}\sum_{n=1}^{H}\left( \hat{\traj}(t_0+n) - {\traj}(t_1+n) \right)^2
% \end{align}
\begin{align}\label{eq: prediction}
{L_{\rm p}} = \frac{1}{H}\sum_{n=1}^{H}\left( \hat{\traj}(t_1+n) - {\traj}(t_1+n) \right)^2.
\end{align}
We will optimize the prediction length $Z(t)=[z_1(t), z_2(t),...,z_I(t)], z_i(t)\leq H$ for each joint in our cross-system design framework, which will be introduced in the next section. 

\subsubsection{Control}
There are different algorithms we can use to control the virtual robotic arm in the Metaverse~\cite{craig2006introduction}. Without loss of generality, we utilize the joint-space position control and \gls{pd} controller~\cite{motion}. Considering the limitations of the control frequency and the processing time, the target angle for each joint will be generated by the control algorithm and subsequently executed for every $N_{\rm c}$ time slots, which is denoted by $\widetilde{\traj}(t) = [\widetilde{\tau_1}(t),..., \widetilde{\tau_I}(t)]$. In the $t_2$-th time slot, for each joint $i$, the target joint position $\widetilde{\tau_i}(t_2)$ to be executed within the $N_{\rm c}$ time slots can be expressed as
\begin{align}\label{Control}
\widetilde{\tau}_{i}(t_2) = k_p\cdot(\hat{\tau}_i(t_1+z_i(t))-\widetilde{\tau}_{i}(t_2-N_c)) + k_d\cdot(\frac{\dif\hat{\tau}_i(t_1+z_i(t))}{\dif t} -\frac{\dif \widetilde{\tau}_i(t_2-N_c)}{\dif t}),
\end{align}
where  $k_p$ and $k_d$ are the proportional and derivative parameters of the \gls{pd} controller, respectively.

\subsubsection{Rendering}In computer graphics, rendering refers to the process of generating controllable and photo-realistic images and videos of virtual scenes~\cite{05849}. In our system, the processing time of each image is denoted by $N_r$ time slots. In other words, the monitor or VR/AR glasses refresh the images at a refresh rate of $1/(N_rt_{\rm s})$ (times/second). 
The relationship between the trajectory of the digital model and the trajectory displayed to the user is given by 
\begin{align}\label{render}
\check{\traj}(t) = F_r(\widetilde{\traj}(t), \theta_r),
\end{align}
where $F_r(\cdot,\theta_r)$ is the rendering function and $\theta_r$ represents the parameters for rendering.

\subsection{KPIs and Communication Load}

 \begin{figure}
\centering
\includegraphics[scale=0.6]{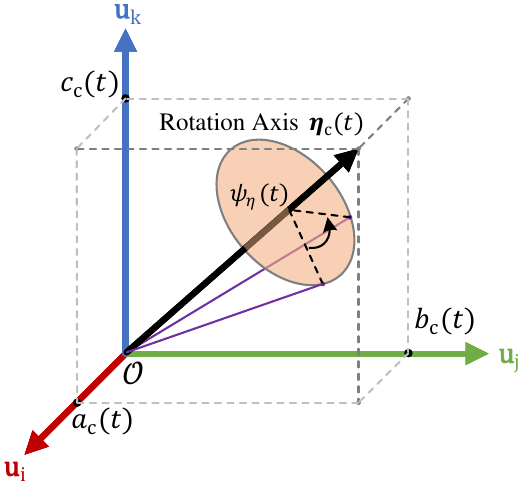}
\caption{Orientation of the end effector to the base coordinate system $\mathcal{O}$.}
\label{fig: quaternion}
\end{figure}

\subsubsection{Task-Oriented KPI}
The end effector of a robotic arm could be a gripper, a drill bit, or a sensor, depending on the specific task. We assume that the real-world end effector has seven degrees of freedom, and the pose of the end effector is denoted by 
\begin{align}
{\mathcal{P}(t)} =[l_\mathrm{{x,r}}(t), l_\mathrm{{y,r}}(t), l_\mathrm{{z,r}}(t), q_\mathrm{{x,r}}(t), q_\mathrm{{y,r}}(t), q_\mathrm{{z,r}}(t), q_\mathrm{{w,r}}(t)].
\end{align} 
Specifically, $l_\mathrm{{x,r}}(t)$, $l_\mathrm{{y,r}}(t)$, $l_\mathrm{{z,r}}(t)$ are the coordinates of the end effector in a three-dimensional Cartesian coordinate system. $[q_\mathrm{{x,r}}(t)$, $q_\mathrm{{y,r}}(t)$, $q_\mathrm{{z,r}}(t)$, $q_\mathrm{{w,r}}(t)]$ is the quaternion representing the orientation of the end effector~\cite{kuipers1999quaternions}.  Quaternions are preferred over other representations, such as Euler angles or rotation matrices, in our context because of their compact representation and their ability to avoid a particular limitation associated with 3D rotation systems, known as gimbal lock, which can cause a loss of degrees of freedom \cite{Diebel_2006_representing}. Please see Appendix \ref{apd: quaternion} for more information. As shown in \cref{fig: quaternion}, the unit vector of the rotation axis,
${\bm{\eta}}_\mathrm{c}(t)=[a_\mathrm{c}(t), b_\mathrm{c}(t), c_\mathrm{c}(t)]$, 
and angle of rotation, $\psi_{\eta}(t)$, can be characterized by $q_\mathrm{{x,r}}(t)$, $q_\mathrm{{y,r}}(t)$, $q_\mathrm{{z,r}}(t)$, and $q_\mathrm{{w,r}}(t)$. In particular, the rotation axis is located in the coordinate system defined by three imaginary unit basic vectors, $\mathbf{u}_\text{i}$, $\mathbf{u}_\text{j}$, and $\mathbf{u}_\text{k}$, which follow special multiplication rules \cite{kuipers1999quaternions}.
The relationship among the quaternions, $\bm{\eta}_\mathrm{c}(t)$, and $\psi_{\eta}(t)$ is expressed by
\begin{align}
    q_\mathrm{{x,r}}(t) &= \sin(\frac{\psi_{\eta}(t)}{2})\cdot a_\mathrm{c}(t) \notag ,\\
    q_\mathrm{{y,r}}(t) &= \sin(\frac{\psi_{\eta}(t)}{2})\cdot b_\mathrm{c}(t) \notag ,\\
    q_\mathrm{{z,r}}(t) &= \sin(\frac{\psi_{\eta}(t)}{2})\cdot c_\mathrm{c}(t) \notag ,\\
    q_\mathrm{{w,r}}(t) &= \cos(\frac{\psi_{\eta}(t)}{2}).
\end{align}
Similarly, the quaternions of the virtual-world robotic arm follow the same rules.

From the $I$ joint angles, ${\mathcal{P}(t)}$ is obtained from the forward kinematics according to
% \begin{align}\label{foward kinemetics}
% [x_r(t), y_r(t), z_r(t), qx_r(t),qw_r(t),qz_r(t)] = F_f[\tau_1(t), ...,\tau_I(t)],
% \end{align}
\begin{align}\label{foward kinemetics}
{\mathcal{P}(t)} = F_f(\traj(t)),
\end{align}
where $F_f(\cdot)$ maps the joint angles to the pose of the end effector (positions and orientations). The expression of \eqref{foward kinemetics} depends on the structure and configuration of the robotic arm. Like ${\mathcal{P}(t)}$, the pose of the end effector displayed to the user also has seven degrees of freedom, denoted by $\check{P}(t) = [l_\mathrm{{x,v}}(t), l_\mathrm{{y,v}}(t), l_\mathrm{{z,v}}(t), q_\mathrm{{x,v}}(t), q_\mathrm{{y,v}}(t), q_\mathrm{{z,v}}(t), q_\mathrm{{w,v}}(t)]$.
A task-oriented KPI is defined as the Euclidean distance between ${\mathcal{P}(t)}$ and $\check{P}(t)$,
\begin{align}\label{weight_norm}
 {\bf{e}}(t) &= \omega_1\cdot\|(l_{\mathrm{x,r}}(t), l_{\mathrm{y,r}}(t), l_\mathrm{{z,r}}(t)), (l_\mathrm{{x,v}}(t), l_\mathrm{{y,v}}(t), l_\mathrm{{z,v}}(t))\|_2 \\ \notag &+\omega_2\cdot\|(q_\mathrm{{x,r}}(t), q_\mathrm{{y,r}}(t), q_\mathrm{{z,r}}(t), q_\mathrm{{w,r}}(t)),(q_\mathrm{{x,v}}(t), q_\mathrm{{y,v}}(t), q_\mathrm{{z,v}}(t), q_\mathrm{{w,v}}(t))\|_2,  
\end{align}
where  $\|\cdot\|_2$ is the $L_2$-norm defined as $\|\cdot\|_2 \triangleq \sqrt{\sum(\cdot)^2}$, and $\omega_1$ and $\omega_2$ are the weighting coefficients. The first term is the position error and the second is the orientation error. Depending on the accuracy requirements of different tasks in the Metaverse,  $\omega_1$ and $\omega_2$ can be set to different values.

\subsubsection{Communication Load}
The \gls{ofdm} communication system is considered in our framework for it is widely deployed in cellular networks. The number of time and frequency resource blocks allocated to a packet is determined by the channel gain and the packet size. We assume that the channel gain and the packet size are two stationary random variables. The average number of resource blocks required in each slot is proportional to the average packet rate. To improve resource utilization efficiency, defined as the average number of resource blocks per slot, we minimize the average number of packets transmitted in each slot.

\section{Constrained Reinforcement Learning for Cross system Design}\label{Sec:CMDP}
\label{sec:kctd3}
In this section, we formulate an optimization problem that minimizes the communication load subject to a constraint on the \gls{cvar} of modeling error by optimizing the scheduling policy and the prediction horizon. In the cross-system design framework, there is no closed-form relationship between the KPIs and the optimization variables. To solve this problem, we develop a \gls{drl} algorithm by integrating domain knowledge into the advanced \gls{ppo} algorithm.

\subsection{Preliminary of PPO}
\gls{ppo} is an advanced reinforcement learning algorithm for solving problems with discrete action space. We chose Proximal Policy Optimization (PPO) as the baseline algorithm due to its simplicity, effectiveness, and high sample efficiency compared to other reinforcement learning algorithms~\cite{2017Proximal}. In addition, PPO maintains a balance between exploration and exploitation and avoids drastic policy updates, which is crucial for managing the complex dynamics of robotics and ensuring stable training~\cite{2017Emergence}. We denote the state and action of \gls{ppo} by ${\bf{s}}_t$ and ${\bf{a}}_t$, respectively. The policy is a mapping from the state to the probabilities of taking different actions, denoted by $\pi_{\theta}({\bf{a}}_t \mid {\bf{s}}_t)$, where $\theta$ are the training parameters of the policy network. With \gls{ppo}, the parameters of the policy are updated according to the following expression,
\begin{align}
\theta_{t+1} = \arg & \mathop {\max }\limits_{\theta} \mathop{\mathbb{E}}\limits_{{\bf{s}}_t,{\bf{a}}_t \sim \pi_\theta}\mathcal{L}({\bf{s}}_t,{\bf{a}}_t,\theta_t,\theta).
\end{align}
The loss function $\mathcal{L}({\bf{s}}_t,{\bf{a}}_t,\theta_t,\theta)$ is given by 
\begin{align}\label{vf new}
\mathcal{L}({\bf{s}}_t,{\bf{a}}_t,\theta_t,\theta) = &\min \left(\frac{\pi_{\theta}({\bf{a}}_t \mid {\bf{s}}_t)}{\pi_{\theta_t}({\bf{a}}_t \mid {\bf{s}}_t)}{A^{\pi_\theta}({\bf{s}}_t,{\bf{a}}_t)},\right. \\ \notag
&\left.\text{clip}\left(\frac{{\pi_\theta}({\bf{a}}_t \mid {\bf{s}}_t)}{\pi_{\theta_t}({\bf{a}}_t \mid {\bf{s}}_t)}, 1-\epsilon, 1+\epsilon,\right)A^{\pi_{\theta_t}}({\bf{s}}_t,{\bf{a}}_t)\right),
\end{align}
where $A({\bf{s}}_t,{\bf{a}}_t)$ is the advantage function defined as the difference between the state-action value function, $Q^{\pi_\theta}({\bf{s}}_t,{\bf{a}}_t)$, and the state value function, $V^{\pi_\theta}({\bf{s}}_t)$, 
\begin{align}\label{eq: advantage}
A^{\pi_\theta}({\bf{s}}_t,{\bf{a}}_t) = Q^{\pi_\theta}({\bf{s}}_t,{\bf{a}}_t) - V^{\pi_\theta}({\bf{s}}_t),
\end{align}
which estimates the advantage of taking action ${\bf{a}}_t$ in state ${\bf{s}}_t$, over other possible actions in the same state~\cite{proximal}. 
In the sequel, we develop our \gls{drl} algorithm by integrating domain knowledge into the \gls{ppo}.

\subsection{Knowledge-Assisted Problem Formulation}\label{Subsec:CMDP}

\subsubsection{State}
The state in the $t$-th time slot consists of two parts: the angles of the $I$ joints and the Jacobian matrix of the real-world robotic arm, i.e., ${{\bf{s}}_t}=\{\traj(t), \mathcal{J}(\traj(t))\}$. In robotics, the Jacobian matrix is critical for analyzing and controlling the motion of robots. It characterizes the relationship between the velocity of the end effector and the velocities of all joints~\cite{craighead2008using}, 
\begin{align}\label{jacobian}
\frac{\Delta{{\mathcal{P}}}(t)}{\Delta t}=\mathcal{J}(\traj(t))\frac{\Delta{\traj}(t)}{\Delta t},
\end{align}
where ${\frac{\Delta{{\mathcal{P}}}(t)}{\Delta t}}$ is the velocity of the end effector, and $\frac{\Delta{\traj}(t)}{\Delta t}$ is the angular velocities of $I$ joints. In the $t$-th time slot, the Jacobian matrix can be obtained from $\traj(t)$ and the kinematic properties of the robotic arm, e.g., \gls{dh} parameters~\cite{craighead2008using}. By multiplying $\Delta t$ on both sides of~\eqref{jacobian}, the relationship between $\Delta{{\mathcal{P}}}(t)$ and $\Delta{\traj}(t)$ is expressed as follows,
\begin{align}\label{jacobian_2}
\Delta{{\mathcal{P}}}(t)=\mathcal{J}(\traj(t))\Delta{\traj}(t),
\end{align}
where $\mathcal{J}(\traj(t))$ shows how sensitive the modeling error of the end effector is to the errors of the $I$ joints. Thus, we take the Jacobian matrix as one part of the state to improve the training efficiency of the \gls{drl} algorithm.

\begin{figure}
\centering
\includegraphics[scale=0.4]{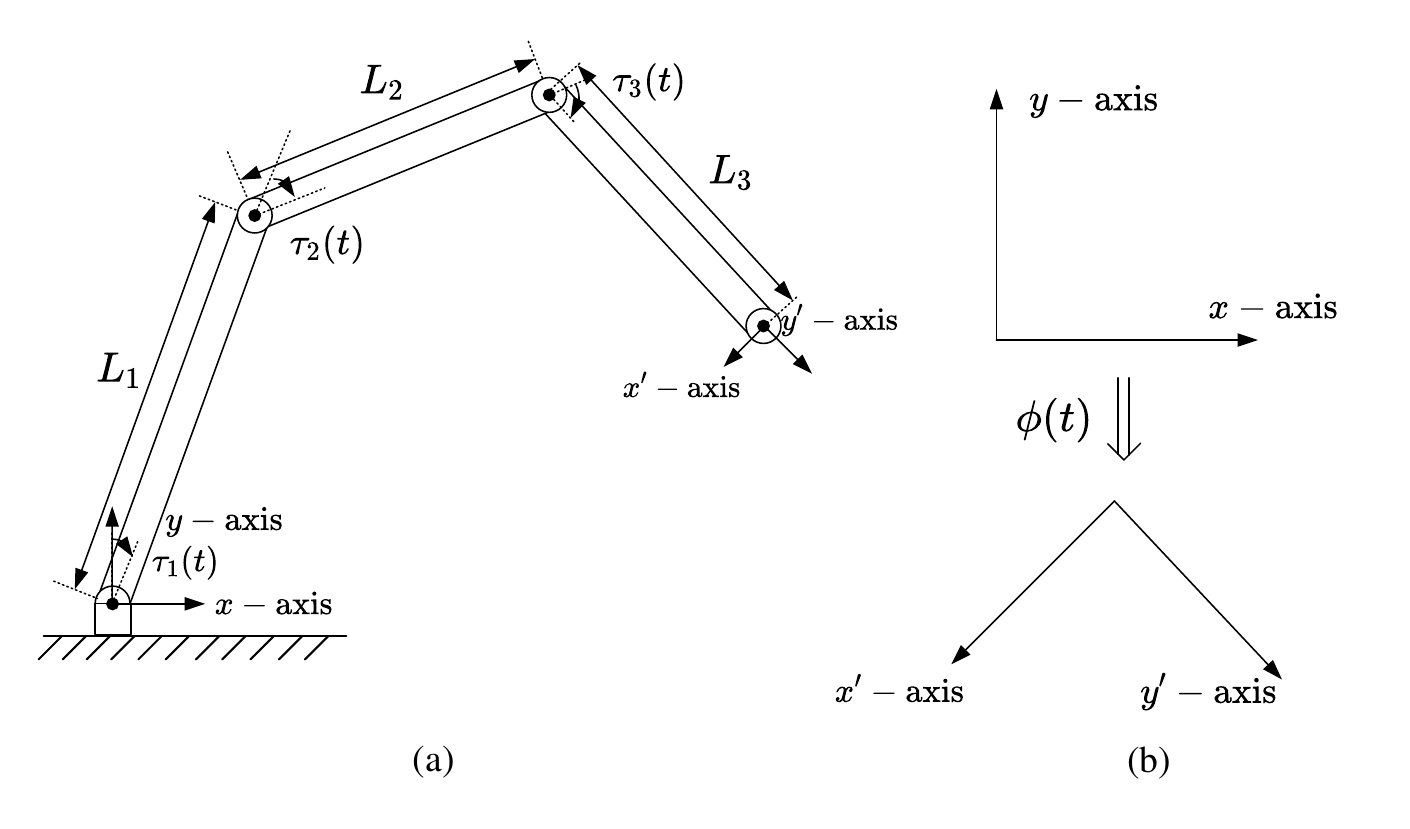}
\caption{Three-link two-dimensional robotic arm model.}
\label{jaco_robot}
\end{figure}

Let's take the three-link two-dimensional robotic arm as an example to show the Jacobian matrix~\cite{craig2006introduction}. With the example in Fig.~\ref{jaco_robot}(a), the forward kinematics in~\eqref{foward kinemetics} can be expressed as follows,
\begin{align}
{\mathcal{P}}(t) =\begin{bmatrix} l_{\mathrm{x,r}}(t) \\ l_{\mathrm{y,r}}(t) \\ \phi(t) \end{bmatrix} =\begin{bmatrix}   
L_1\cdot\cos{\tau_1(t)}+L_2\cdot\cos{(\tau_1(t)+\tau_2(t))}+L_3\cdot\cos{(\tau_1(t)+\tau_2(t)+\tau_3(t))}\\
L_1\cdot\sin{\tau_1(t)}+L_2\cdot\sin{(\tau_1(t)+\tau_2(t))}+L_3\cdot\sin{(\tau_1(t)+\tau_2(t)+\tau_3(t))}\\
\tau_1(t)+\tau_2(t)+\tau_3(t)
\end{bmatrix},
\end{align}
where $L_1$, $L_2$, and $L_3$ are the lengths of the three links, respectively. As shown in Fig.~\ref{jaco_robot}(b),  $\phi(t)$ is the angle between $x$-axis and $x'$-axis in the clockwise direction. Then, the Jacobian matrix can be obtained by 

\begin{align}
        \mathcal{J}(\traj(t))= \begin{bmatrix}   
        \frac{\partial{l_{\mathrm{x,r}}(t)}}{\partial{\tau_1(t)}}    & \frac{\partial{l_{\mathrm{x,r}}(t)}}{\partial{\tau_2(t)}}     & \frac{\partial{l_{\mathrm{x,r}}(t)}}{\partial{\tau_3(t)}}\\
        \frac{\partial{ l_{\mathrm{y,r}}(t)}}{\partial{\tau_1(t)}}    & \frac{\partial{ l_{\mathrm{y,r}}(t)}}{\partial{\tau_2(t)}}     & \frac{\partial{ l_{\mathrm{y,r}}(t)}}{\partial{\tau_3(t)}}\\
        \frac{\partial{\phi}}{\partial{\tau_1(t)}} & \frac{\partial{\phi}}{\partial{\tau_2(t)}}  & \frac{\partial{\phi}}{\partial{\tau_3(t)}} \end{bmatrix}
\end{align}

where $\mathcal{J}(\traj(t))$ consists of all partial derivatives of ${\mathcal{P}}(t)$. Specifically, the first two rows of the matrix are related to the partial derivatives of the position coordinates, while the last row is related to the partial derivatives of the angle of the end effector which is shown in Fig.~\ref{jaco_robot}(b). Thus, each element is calculated by

\begin{align}
    \frac{\partial{{l_{\mathrm{x,r}}(t)}}}{\partial{\tau_1(t)}} &= -L_1\cdot\sin{\tau_1(t)} - L_2\cdot\sin{(\tau_1(t) + \tau_2(t))} - L_3\cdot\sin{(\tau_1(t) + \tau_2(t) + \tau_3(t))}, \\
    \frac{\partial{{l_{\mathrm{x,r}}(t)}}}{\partial{\tau_2(t)}} &= - L_2\cdot\sin{(\tau_1(t) + \tau_2(t))} - L_3\cdot\sin{(\tau_1(t) + \tau_2(t) + \tau_3(t))},
    \\
    \frac{\partial{{l_{\mathrm{x,r}}(t)}}}{\partial{\tau_3(t)}} &= - L_3\cdot\sin{(\tau_1(t) + \tau_2(t) + \tau_3(t))}, \\              
   \frac{\partial{{l_{\mathrm{y,r}}(t)}}}{\partial{\tau_1(t)}} &= -L_1\cdot\sin{\tau_1(t)} - L_2\cdot\sin{(\tau_1(t) + \tau_2(t))} - L_3\cdot\sin{(\tau_1(t) + \tau_2(t) + \tau_3(t))}, \\
    \frac{\partial{{l_{\mathrm{y,r}}(t)}}}{\partial{\tau_2(t)}} &= - L_2\cdot\sin{(\tau_1(t) + \tau_2(t))} - L_3\cdot\sin{(\tau_1(t) + \tau_2(t) + \tau_3(t))}, \\
    \frac{\partial{{l_{\mathrm{y,r}}(t)}}}{\partial{\tau_3(t)}} &= - L_3\cdot\sin{(\tau_1(t) + \tau_2(t) + \tau_3(t))}, \\
    \frac{\partial{\phi}}{\partial{\tau_1(t)}} &= 1, \qquad \frac{\partial{\phi}}{\partial{\tau_2(t)}} = 1, \qquad \frac{\partial{\phi}}{\partial{\tau_3(t)}} = 1.
\end{align}

From the above description, we can see that the modeling error of the end effector is more sensitive to the error of the joint that is far away from the end effector and less sensitive to the error of the joint that is close to the end effector. The robotic arm in our prototype has more than three joints and can move in a three-dimensional space. Thus, the Jacobian matrix could be more complicated than the two-dimensional robotic arm in Fig.~\ref{jaco_robot}.

The state, including joint angles and elements of the Jacobian matrix, needs to be normalized 
before feeding it into the neural network. We first find the maximum and minimum values of each joint angle and each element of the Jacobian matrix from the data set. Then, these values are employed to normalize the state within the range of $(0,1)$.

\subsubsection{Action}
The action to be taken in the $t$-th time slot includes the joints to be scheduled, $X(t)$, and the optimal prediction horizons $Z(t)$. Although the prediction horizon needs to be transmitted to the server, $Z(t)$ is an integer ranging from $1$ to $500$. Thus, the communication overhead for updating $Z(t)$ is negligible compared to the update of the joint angle with high precision. We denote the action by ${{\bf{a}}_t} = [{{\bf{a}}^{[1]}_t},{{\bf{a}}^{[2]}_t}] = [X(t), Z(t)]$, where the action of the $i$-th joint is denoted by ${{\bf{a}}^{[1]}_{t,i}} = x_i(t)$ and ${{\bf{a}}^{[2]}_{t,i}} = z_i(t)$.

\subsubsection{Instantaneous Reward and Cost} % \textcolor{red}{The expressions of instantaneous and long-term rewards}.
Given the state and the action taken in the $t$-th time slot, the instantaneous reward, denoted by $r({\bf{s}}(t), {\bf{a}}(t))$, is the communication load reduction compared with a benchmark that all joints are scheduled in every time slot. The instantaneous cost $c({\bf{s}}(t), {\bf{a}}(t))$ is set to ${\bf{e}}(t)$ in \eqref{weight_norm}. 

\subsubsection{Policy} 
\begin{figure}
      \centering
      \includegraphics[scale=0.40]{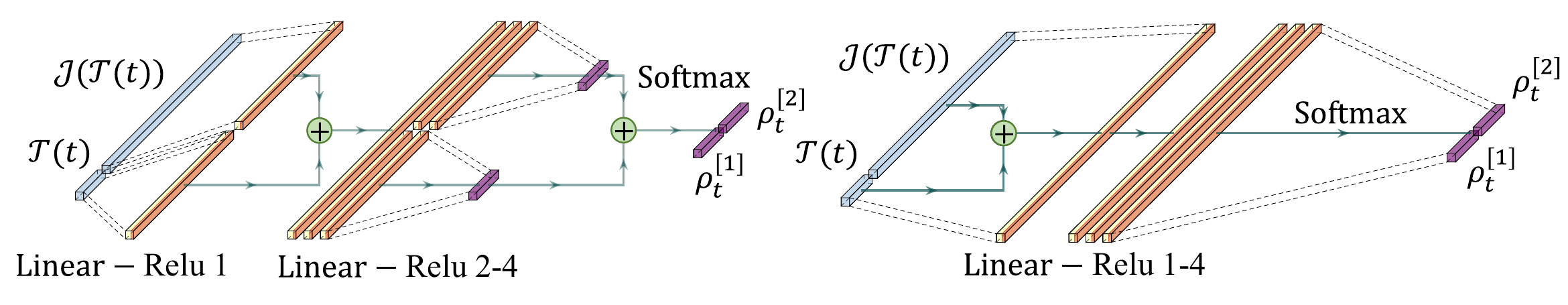} 
      \caption{Structures of neural networks: Two-branch neural network and fully-connected neural network.}
      \label{Cascade neural network}
  \end{figure}

The policy is represented by a neural network, $\pi_{\theta}\left( {\bf{s}}_t \right)$, where $\theta$ represents the training parameters. The network consists of multiple fully connected layers as shown in Fig.~\ref{Cascade neural network}. In our study, the inputs to the policy networks include two different states: the angles of joints and the Jacobian matrix of the real-world robotic arm. The raw data of a joint angle has complete information and requires a complex neural network for feature extraction. The Jacobi matrix provides information that has been processed based on domain knowledge, and we can use a simple neural network for feature extraction. To handle different types of input information, we designed the two-branch neural network. Meanwhile, the two branches are designed to optimize two types of actions separately, i.e., the scheduling of a joint and the prediction horizon.

Specifically, the first two layers are designed for feature extraction, where the input denoted by $\{\traj(t), \mathcal{J}(\traj(t))\}$ is transformed into a more compact and informative representation that captures the underlying patterns. Then, we decouple the neural network into two parallel branches. The first branch is for the scheduling policy, $\pi_\theta^{[1]}$, and the second branch is for the policy of optimizing of the prediction horizons, $\pi_\theta^{[2]}$. After that, two branches are concatenated in the final linear layer. Followed by the Softmax activation function, the distribution of two actions, i.e., $\rho^{[1]}_t$ and $\rho^{[2]}_t$ are generated. Finally, two actions, i.e., ${{\bf{a}}^{[1]}_t}$ and ${{\bf{a}}^{[2]}_t}$ are sampled from $\rho^{[1]}_t$ and $\rho^{[2]}_t$, respectively.

Specifically, $\pi_\theta^{[1]}$ maps the state 
${\bf{s}}_t$ to the distribution of ${{\bf{a}}^{[1]}_{t,i}}$, denoted by $\rho^{[1]}_t \in \mathbb{R}^{2 \times I}$. The $i$-th column of $\rho^{[1]}_t$ is defined as follows,
\begin{align}\label{eq:sampling}
{\rho^{[1]}_{t,i}} \triangleq
\begin{pmatrix}
   \Pr\{a^{[1]}_{t,i}=1\}\\
   \Pr\{a^{[1]}_{t,i}=0\}
\end{pmatrix}.
\end{align}
Similarly, $\pi_\theta^{[2]}$ maps the state ${\bf{s}}_t$ to the distribution of ${{\bf{a}}^{[2]}_{t,i}}$, denoted by $\rho^{[2]}_t \in \mathbb{R}^{H \times I}$. The $i$-th column of $\rho^{[2]}_t$ is defined as follows,
\begin{align}\label{eq:prediction}
{\rho^{[2]}_{t,i}} \triangleq \left[\Pr\{a^{[2]}_{t,i}=1\}, \Pr\{a^{[2]}_{t,i}=2\}, ..., \Pr\{a^{[2]}_{t,i}=H\}\right]^{\rm T}.
\end{align}
 Once the distributions are obtained, ${{\bf{a}}^{[1]}_{t,i}}$ and ${{\bf{a}}^{[2]}_{t,i}}$ can be sampled from \eqref{eq:sampling} and \eqref{eq:prediction}, respectively. Here, the probability of each action being sampled is based on the weight located in the corresponding elements~\cite{multinomial}. The policies of different joints are represented by DNNs with the same structure. If there are more joints and sensors, they can reuse the DNN and fine-tune the parameters with few-shot learning. In this way, we can address the scalability issue.

\subsubsection{Long-Term Reward and Cost}
Following the policy $\pi_{\theta}$, the long-term reward and long-term cost are defined as $R^{\pi_{\theta}} = {\mathop{\mathbb{E}}}[\sum\nolimits_{t = 0}^\infty  {{\gamma ^t}} r({\bf{s}}(t), {\bf{a}}(t)]$ and $C^{\pi_{\theta}} = {\mathop{\mathbb{E}}}[\sum\nolimits_{t = 0}^\infty  {{\gamma ^t}} c({\bf{s}}(t), {\bf{a}}(t)]$, respectively, where $\gamma$ is the discount factor~\cite{xu2021crpo}. To estimate the long-term reward and long-term cost, we can use the state-value function or the state-action value function. The state-value function and the state-action-value function of the long-term reward are defined as
\begin{align}
   &V_r^{\pi_{\theta}}({\bf{s}}) = {\mathop{\mathbb{E}}}[\sum\nolimits_{t = 0}^\infty  {{ \gamma ^t}}r({\bf{s}}(t), {\bf{a}}(t)) \mid {\bf{s}}_0={\bf{s}},\ \pi_{\theta}],\\
   &Q_r^{\pi_{\theta}}({\bf{s}}, {\bf{a}}) = {\mathop{\mathbb{E}}}[\sum\nolimits_{t = 0}^\infty  {{ \gamma ^t}}r({\bf{s}}(t), {\bf{a}}(t)) \mid {\bf{s}}_0={\bf{s}},\ {\bf{a}}_0={\bf{a}},\ \pi_{\theta}],
\end{align}
respectively. The advantage function is given by $A^{\pi_{\theta}}_r ({\bf{s}}, {\bf{a}})  =  Q_r^{\pi_{\theta}}({\bf{s}}, {\bf{a}}) - V_r^{\pi_{\theta}}({\bf{s}})$.
Like the long-term reward, the state-value function and the state-action-value function of the long-term cost are defined as 
\begin{align}
   &V_c^{\pi_{\theta}}({\bf{s}}) = {\mathop{\mathbb{E}}}[\sum\nolimits_{t = 0}^\infty  {{ \gamma ^t}}c({\bf{s}}(t), {\bf{a}}(t)) \mid {\bf{s}}_0={\bf{s}},\ \pi_{\theta}],\\
   &Q_c^{\pi_{\theta}}({\bf{s}}, {\bf{a}}) = {\mathop{\mathbb{E}}}[\sum\nolimits_{t = 0}^\infty  {{ \gamma ^t}}c({\bf{s}}(t), {\bf{a}}(t)) \mid {\bf{s}}_0={\bf{s}},\ {\bf{a}}_0={\bf{a}},\ \pi_{\theta}],
\end{align}
respectively. The advantage function is given by $A^{\pi_{\theta}}_c({\bf{s}}, {\bf{a}}) =  Q_c^{\pi_{\theta}}({\bf{s}}, {\bf{a}}) - V_c^{\pi_{\theta}}({\bf{s}})$.

\subsubsection{Modeling Accuracy Constraint} 
To guarantee the long-term modeling accuracy constraint, a straightforward approach is to evaluate $C^{\pi_{\theta}}$ by using $V_c^{\pi_{\theta}}({\bf{s}})$ or $Q_c^{\pi_{\theta}}({\bf{s}}, {\bf{a}})$. Note that the average long-term cost may not be applicable for mission-critical tasks in the Metaverse. For example, in haptic communications, users cannot recognize errors below a certain threshold, known as Just Noticeable Difference \cite{feyzabadi2013human}. For mission-critical tasks, we propose to use ~\gls{cvar} of $Q_c^{\pi_{\theta}}({\bf{s}}_t, {\bf{a}}_t)$ as the \gls{kpi}. \gls{cvar} is a well-known risk measure used in financial portfolio analysis that depicts the cost in the tail of the risk distribution~\cite{01629}. 
The expression of \gls{cvar} of $Q_c^{\pi_{\theta}}({\bf{s}}_t, {\bf{a}}_t)$ is given by
 \begin{align}\label{CVaR}
\text{CVaR}_\alpha[Q_c^{\pi_{\theta}}({\bf{s}}_t, {\bf{a}}_t)] = \min\limits_{ v \in \mathcal{R}}  \left(  v + \frac{1}{1-\alpha} \mathop{\mathbb{E}}\limits_{{\bf{s}}_t,{\bf{a}}_t \sim \pi_\theta}
\left\{ [Q_c^{\pi_{\theta}}({\bf{s}}_t, {\bf{a}}_t)- v]^{+}\right\}\right),
\end{align}
where $(x)^{+} = \text{max}(x,0)$. $\alpha \in (0,1)$ is the confidence level, and 
$Q_c^{\pi_{\theta}}({\bf{s}}_t, {\bf{a}}_t)$ is equal to the average of the worst-case $\alpha$-fraction of losses under optimal conditions~\cite{rockafellar2000optimization}.

\subsubsection{Problem Formulation} The goal is to find the optimal policy ${\pi_{\theta}^{*}}$ that maximizes the long-term reward $R^{\pi_{\theta}}$ subject to the constraint on \gls{cvar} of the long-term cost $C^{\pi_{\theta}}$. Thus, the problem can be formulated as follows:
\begin{align}
  {\pi_{\theta}^{*}} = &\arg \mathop {\max }\limits_{\theta} Q_r^{\pi_{\theta}}({\bf{s}}_t, {\bf{a}}_t)\hfill \label{tt}\\
  &\,\text{s.t.}
\quad \ \; \text{CVaR}_\alpha[Q_c^{\pi_{\theta}}({\bf{s}}_t, {\bf{a}}_t)] \le {\frac{{{\Gamma _c}}}{{1 - \gamma }}} \hfill \tag{\ref{tt}{a}} \label{tta},
\end{align}
where ${\Gamma_c}$ is the modeling error depending on the requirements of specific tasks in the Metaverse. 

\begin{algorithm}[t]\label{alg.l1}
\begin{algorithmic}[1]
\renewcommand{\algorithmicrequire}{\textbf{Input:}} %Use Input in the format of Algorithm
\renewcommand{\algorithmicensure}{\textbf{Output:}} %Use Output in the format of Algorithm

\caption{C-PPO} 
\REQUIRE
initial the parameters of neural network including policy network $\theta_0$, initial state $\mathbf{s}_{0}$, step length $\beta$ 
\FOR{$t=0,1,2,\dots,T-1$}
    \STATE Observe $\mathbf{s}_{t}$ and generate action from current policy $\pi_{\theta_t}({[\mathbf{a}_{t}^{[1]}, \mathbf{a}_{t}^{[2]}] \mid \mathbf{s}_{t})}$ 
    \STATE Transmit the packets based on action $\mathbf{a}_{t}^{[1]}$
    \STATE Reconstruct the trajectory based on received packets by (\ref{reconstruct function})
    \STATE Predict the trajectory based on action $\mathbf{a}_{t}^{[2]}$ by (\ref{prediction})
    \STATE Store state $\mathbf{s}_t$, action $\mathbf{a}_t$, reward $r_{t}$, cost $c_{t}$, and next state $\mathbf{s}_{t+1}$
    \FOR{$k= 1,2,\dots,K$}
        \STATE Update   $\text{CVaR}_\alpha[Q_c^{\pi_{\theta}}({\bf{s}}_t, {\bf{a}}_t)]$ by \eqref{gradient descent}
    \ENDFOR
    \STATE Compute the advantage function ${A_c^{\pi_\theta}({\bf{s}}_t,{\bf{a}}_t)}$ and ${A_r^{\pi_\theta}({\bf{s}}_t,{\bf{a}}_t)}$ based on~\eqref{vf new},~\eqref{vf cost}
\IF{$\text{CVaR}_\alpha[Q_c^{\pi_{\theta}}({\bf{s}}_t, {\bf{a}}_t)] \le \frac{\Gamma_c}{1-\gamma} $}
    \STATE Take one-step policy update towards maximizing $\mathcal{L}_r({\bf{s}}_t,{\bf{a}}_t,\theta_t,\theta)$ : $\theta_t \to \theta_{t+1}$
\ELSE
    \STATE Take one-step policy update towards maximizing $\mathcal{L}_c({\bf{s}}_t,{\bf{a}}_t,\theta_t,\theta)$: $\theta_t \to \theta_{t+1}$
\ENDIF
\ENDFOR
\ENSURE
Optimal policy ${\pi_{\theta}^{*}}$
\end{algorithmic}
\end{algorithm}

\subsection{C-PPO Algorithm}
To guarantee the modeling accuracy constraint, we develop a \gls{cppo} algorithm by integrating \gls{ppo} and \gls{cvar} into the \gls{crpo} algorithm, which is a safe reinforcement learning algorithm with convergence guarantee~\cite{xu2021crpo}. The basic idea of the \gls{crpo} algorithm is to maximize the long-term reward when the constraint is satisfied and minimize the long-term cost when the constraint is violated. 

The details of the \gls{cppo} algorithm can be found in Algorithm~1. In the $t$-th step, we first update the  threshold of \gls{cvar} according to the current policy by the gradient descent,
 \begin{align}\label{gradient descent}
v^{(k+1)} =v^{(k)}-\beta\frac{\Delta\text{CVaR}_\alpha(v^{(k)})}{\Delta v^{(k)}},
\end{align}
where $\beta$ is the learning rate and it takes $K$ steps of gradient descent to update the threshold, $k=1,...,K$. Then, we validate whether the constraint can be satisfied or not. If the constraint in \eqref{tta} is satisfied, we maximize $\mathcal{L}_r({\bf{s}}_t,{\bf{a}}_t,\theta_t,\theta)$ which is obtained by substituting $A^{\pi_{\theta}}_r ({\bf{s}}_t, {\bf{a}}_t)$ into \eqref{vf new}. Otherwise, we minimize $\mathcal{L}_c({\bf{s}}_t,{\bf{a}}_t,\theta_t,\theta)$ defined as follows, 
 \begin{align}\label{vf cost}
\mathcal{L}_c({\bf{s}}_t,{\bf{a}}_t,\theta_t,\theta) = &\min \left(\frac{\pi_{\theta}({\bf{a}}_t \mid {\bf{s}}_t)}{\pi_{\theta_t}({\bf{a}}_t \mid {\bf{s}}_t)}{A_c^{\pi_\theta}({\bf{s}}_t,{\bf{a}}_t)},\right. \\\notag
&\left.\text{clip}\left(\frac{{\pi_\theta}({\bf{a}}_t \mid {\bf{s}}_t)}{\pi_{\theta_t}({\bf{a}}_t \mid {\bf{s}}_t)}, 1-\epsilon, 1+\epsilon,\right)A_c^{\pi_{\theta_t}}({\bf{s}}_t,{\bf{a}}_t)\right),
\end{align}
where ${A_c^{\pi_\theta}({\bf{s}}_t,{\bf{a}}_t)}$ is obtained by \gls{gae}~\cite{proximal, stable-baselines3}. With \gls{cppo}, the parameters are updated according to the following expression,
 \begin{align}\label{pu_s}
\theta_{t+1} = \begin{cases} \theta_{t} + \alpha\nabla_\theta\mathcal{L}_r({\bf{s}}_t,{\bf{a}}_t,\theta_t,\theta ), & {\text{CVaR}_\alpha}[Q_c^{\pi_{\theta}}({\bf{s}}_t, {\bf{a}}_t)] \le \frac{\Gamma_c}{1-\gamma}, \\
\theta_{t} - \alpha\nabla_\theta\mathcal{L}_c({\bf{s}}_t,{\bf{a}}_t,\theta_t,\theta ), & {\text{CVaR}_\alpha}[Q_c^{\pi_{\theta}}({\bf{s}}_t, {\bf{a}}_t)] > \frac{\Gamma_c}{1-\gamma}. \\
\end{cases}
\end{align}
It is worth noting that the policy gradient and the \gls{cvar} gradient can be updated with different learning rates. To guarantee a stable \gls{cvar} constraint when performing the policy gradient, we update the threshold of \gls{cvar} with a higher frequency. 

\section{Prototype Design and Performance evaluation}
\label{sec:results}

\begin{figure}
            \centering
            \includegraphics[scale=0.40]{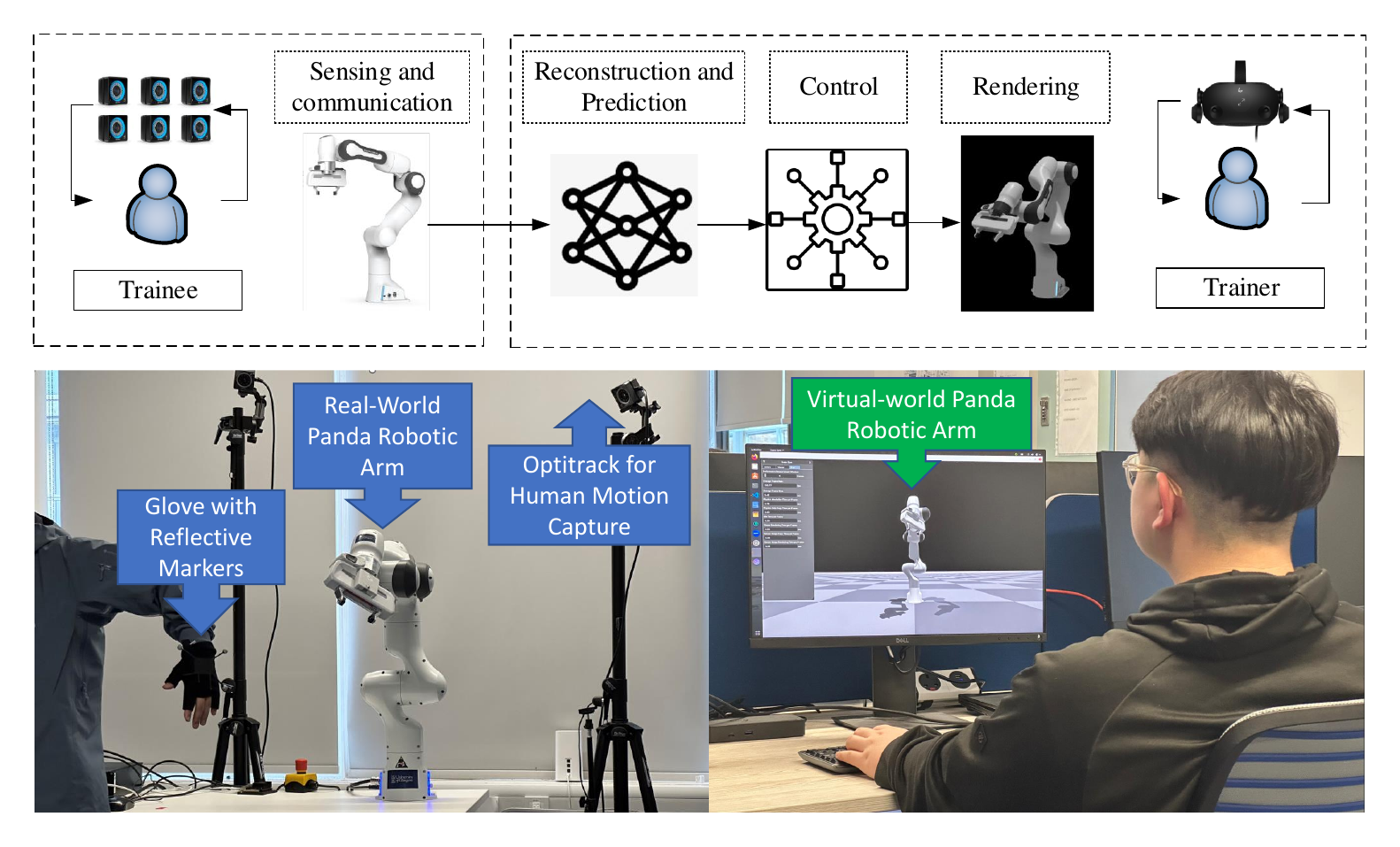}
           \caption{Illustration of our prototype system. 1) Bottom left photo: A real-world robotic arm is controlled by a human trainee, 2) Bottom right photo: A digital model of the robotic arm in the Metaverse is rendered and presented to a trainer, 3) The diagram on the top of the two photos shows the system functions implemented at the two sides.}
           \label{fig: prototype}
  \end{figure}

In this section, we demonstrate our cross-system prototype\footnote{We will release our dataset and source code of C-PPO if this paper gets published. } design as shown in Fig.~\ref{fig: prototype}. Based on the prototype, we first evaluate the effectiveness of the proposed cross-system design framework and then compare the performance with different benchmarks.

\subsection{Prototype Design}
\subsubsection{Real-World Robotic Arm}
We adopt an industrial-grade robotic arm system, Franka Emika Panda~\cite{libfranka}, in our prototype. The robotic arm has seven \gls{dof} and can achieve up to $2$~m/s end effector speed and $\pm0.1$~mm repeatability. In our prototype design, we use five \gls{dof} of the real-world robotic arm. The trainer wearing optical markers controls the robotic arm via the state-of-the-art motion capture system with six motion cameras, OptiTrack Prime-13 \cite{web_optitrack}.
The reason why we use OptiTrack is because 1) It is a real-time motion capture system with high accuracy and low latency. This real-time high-fidelity tracking is essential for maintaining an accurate digital twin in the Metaverse and for evaluating the performance of our task-oriented, cross-system design framework. The use of OptiTrack also allows us to quantify the task-oriented KPI, i.e., average tracking error between the real-world robotic arm and its digital model in the Metaverse. 2) The OptiTrack motion tracking system can be scaled to accommodate various tracking volumes, from small studio setups to large outdoor environments. This scalability makes it versatile and adaptable for different types of motion capture objects in the Metaverse. 3) Our proposed framework is not limited to the use of OptiTrack as a motion capture device. Flexible positioning and motion tracking system can be used bsaed on the practical accuracy demands of the Metaverse applications.

The robotic arm receives the target pose from the motion capture system and then maps the pose to joint angles. After that, the robotic arm applies a {proportional-integral-derivative} method~\cite{ang2005pid} for control, which converts the joint angles to a series of commands. Meanwhile, built-in sensors in the robotic arm measure joint angles, angular velocities, and inertial torque of each joint~\cite{libfranka}.

\subsubsection{Virtual Robotic Arm in the Metaverse}
We establish the Metaverse in the Nvidia Isaac Gym~\cite{gym}, a cutting-edge robotics simulation engine that uses state-of-the-art algorithms and physics engines to simulate the movement and behavior of robots in various environments. Meanwhile, we simulate the communication system between the real-world robotic arm and the Nvidia Isaac Gym by introducing a Gaussian-distributed communication latency. Its mean and the standard deviation are set to $10$~ms and $1$~ms, respectively.

\begin{table}[t]\label{parameters}
\renewcommand{\arraystretch}{0.7}
\centering
\caption{System Parameters for Performance Evaluation}
\begin{threeparttable} 

\begin{tabular}{|l|l|l|l|}
\hline 
\multicolumn{2}{|c|}{\textbf{Prototype Setup}}  & \multicolumn{2}{c|}{\textbf{Learning Setup}} \\ 
\hline
\bf{Parameters} & \bf{Values} &\bf{Parameters} & \bf{Values}\\
\hline
Duration of time slot& \SI{1}{\milli\second} 
& Learning rate of actor network & $\num{3e-4}$\\
\hline
Number of joints $I$ & $5$ 
& Learning rate of critic network & $\num{3e-4}$\\
\hline
Input length of reconstruction function $W_l$ & \SI{2000}{\milli\second} 
& Learning rate of cost network & $\num{3e-4}$\\
\hline
Input length of prediction function $W_p$ & \SI{2000}{\milli\second} 
& Batch size& {256} \\
\hline
Prediction horizon of prediction function $H$ & \SI{500}{\milli\second} 
&Discount factor $\gamma$ & $\num{0.99}$\\
\hline
Control interval $N_c$& \SI{2}{\milli\second}
& Clip ratio in the loss functions of C-PPO $\epsilon$ & 0.2\\
\hline
Refresh rate of image $1/{N_r}{t_s}$ & \SI{60}{Hz}
&Total steps for updating CVaR $K$ & 500\\
\hline
Experimental time & \SI{5e4}{\milli\second}
&Learning rate of constraint $\beta$ & $\num{2e-3}$ \\
\hline
Weighting coefficient of position $\omega_1$ & {0.5}
&Confidence level of CVaR $1- \alpha_c$ & 0.95\\
\hline 
Weighting coefficient of orientation $\omega_2$ & {0.5}
&  & \\
\hline
\end{tabular}

\end{threeparttable}
\label{tab:sys_param}
\end{table}

\subsection{System Setup}
\subsubsection{Parameters of the Prototype} 
For the prototype design, five joint angles of the real-world robotic arm are controlled by the trainee and measured by built-in sensors in each time slot. 
The measured data of the real-world and virtual-world robotic arms are recorded in the CSV format file. For the training process of the predictor, \textit{informer}, we set the prediction input length $W_p$ to $ 2000\ \text{ms}$ and output length $H $ to $ 500\ \text{ms}$. In Nvidia Issac Gym, we set the control interval $N_c$ to $ 2\ \text{ms}$. The frequency at which the robotic arm interacts with the virtual environment is $500$~Hz. The method for calculating the Jacobian Matrix can be found in Appendix \ref{app:A}. The default parameters of the prototype and learning algorithm are listed in Table~\ref{tab:sys_param}, unless otherwise specified.

\subsubsection{Benchmarks} We build our \gls{cppo} algorithm and four benchmarks in the well-known \gls{drl} library \textit{Stable-Baselines3}~\cite{stable-baselines3}. The legends of the benchmarks are ``W2B", ``WJM", ``WCVaR", and ``WDK", respectively. (a) In W2B, the two-branch neural network is replaced with a fully-connected neural network; (b) In WJM, the Jacobian matrix is not included in the state; (c) In WCVaR, the \gls{cvar} of the modeling error is replaced with the average modeling error in the constraint; (d) WDK is a simplified \gls{cppo} without using any domain knowledge (i.e., the two-branch neural network, Jacobian matrix, and \gls{cvar} of the modeling error). With the above benchmarks, we will illustrate the impact of different types of domain knowledge on the performance of our \gls{cppo} algorithm.

\begin{figure}

  \subfigure[\ \ \ Average packet rate in \\ each \ episode.]{%caption of the subfigure
  \label{10_reward}%%label for first subfigure
  \includegraphics[scale=0.189]{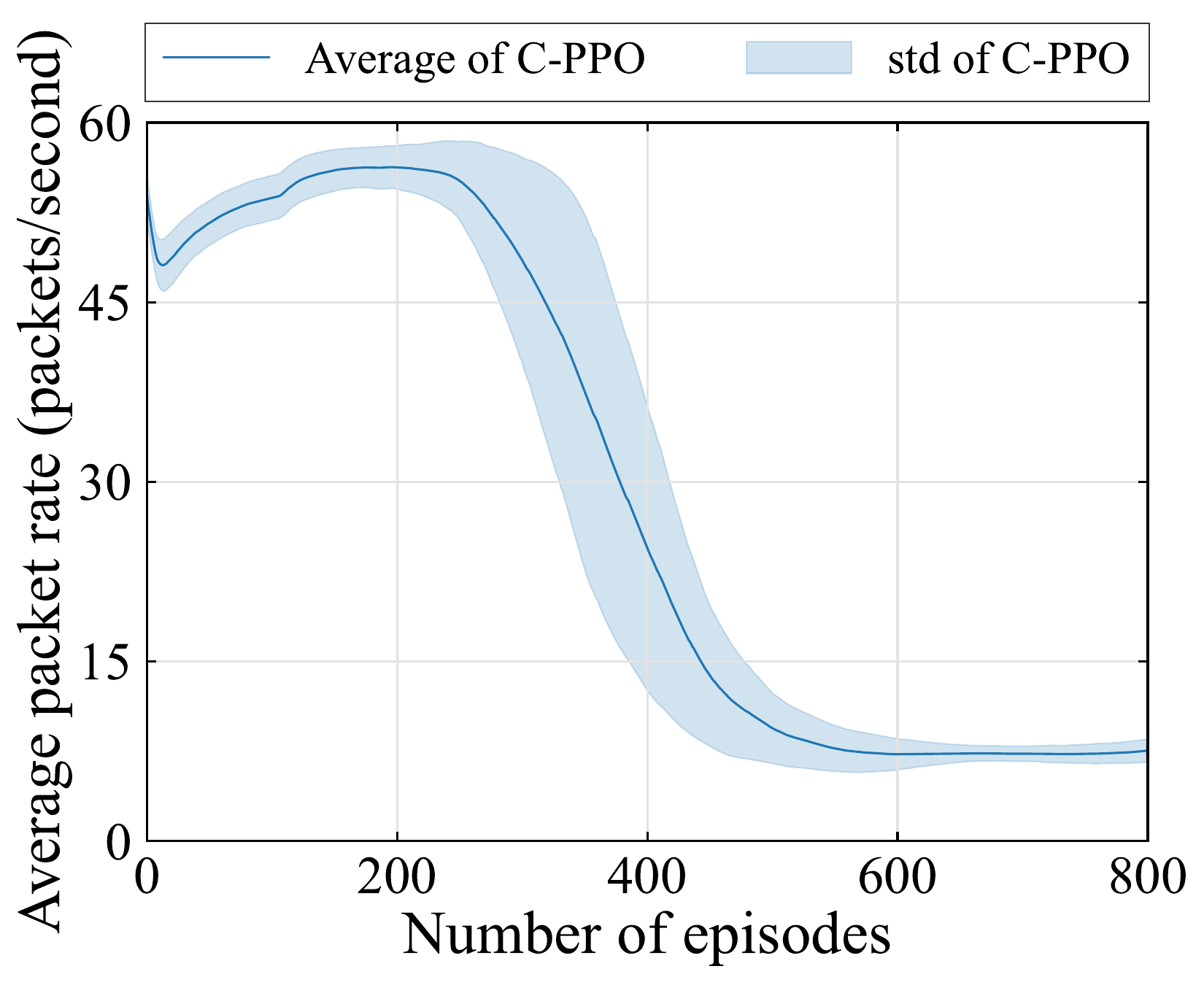}
  }
  \subfigure[\ \ Average modeling error \\ \ \ \ \ in each episode.]{%caption of the subfigure
  \label{10_cost}%%label for second subfigure
  \includegraphics[scale=0.189]{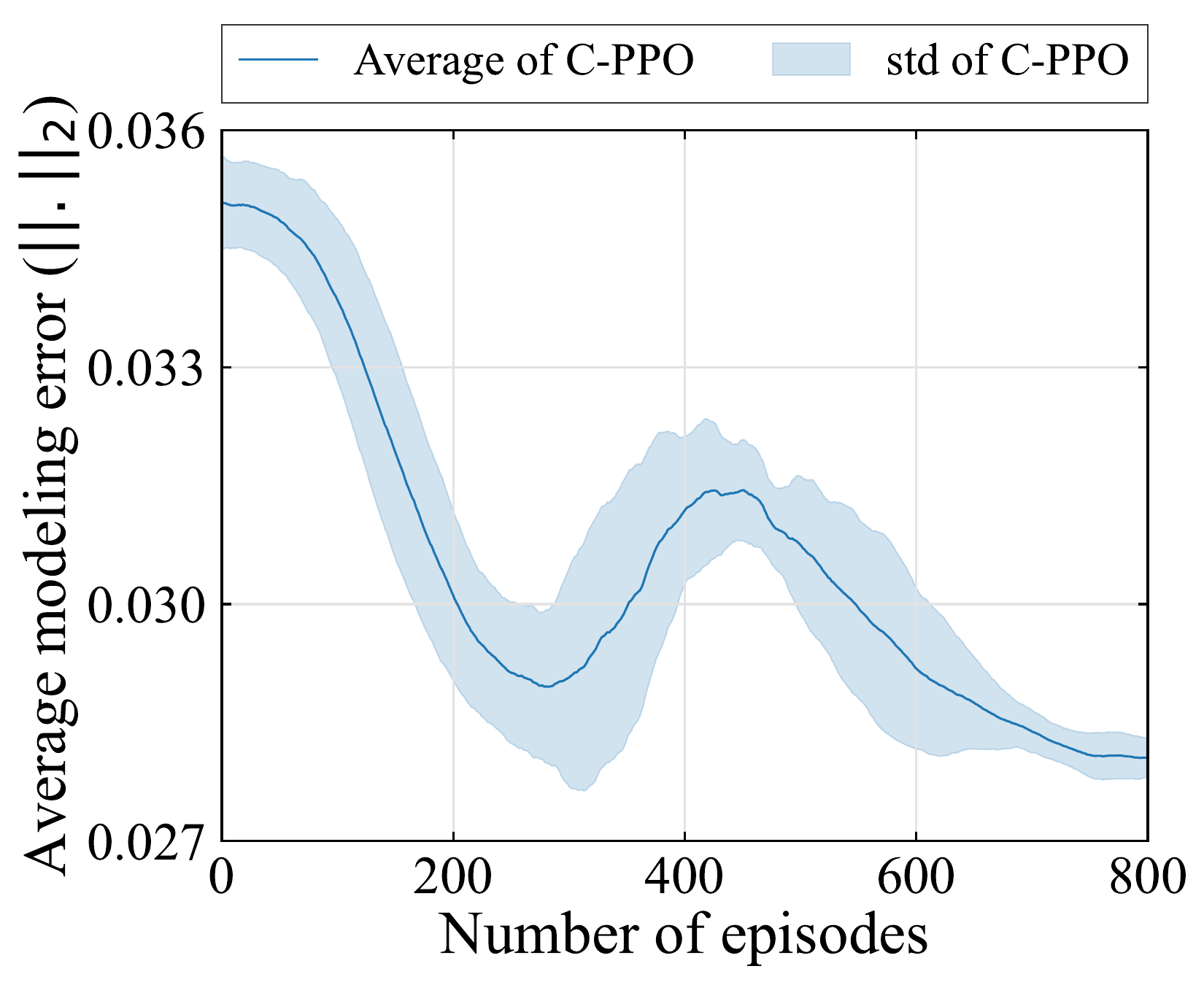}}
  \subfigure[\ \ Average value of \gls{cvar} \\ \ \ \ \ in each episode.]{%caption of the subfigure
  \label{10_cvar}%%label for second subfigure
  \includegraphics[scale=0.189]{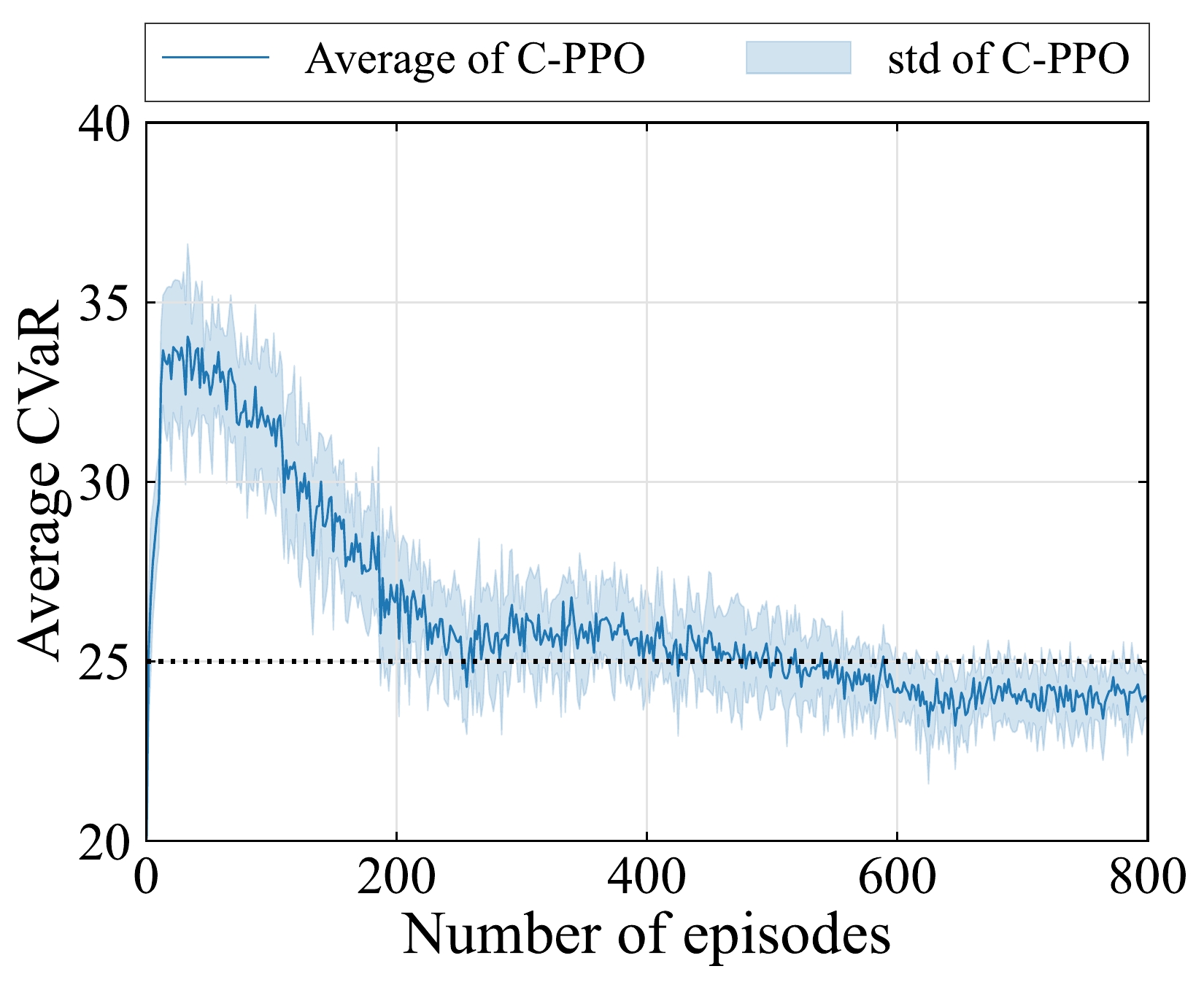}}  
  \caption{Performance evaluation.} % in the training stage}
  \label{fig: 10_times_evaluation}%% label for entire figure
\end{figure}

\subsection{Evaluation of C-PPO Algorithm}
As shown in Fig.~\ref{fig: 10_times_evaluation}, the performance of \gls{cppo} is evaluated by average packet rate, average modeling error, and average \gls{cvar}. We train \gls{cppo} with 800 episodes to show the trends in performance changes. The average packet rate at the start point is defined as a baseline. The average packet rate reaches 8 packets/second, which saves 83.3\% communication load than the baseline.  In addition, optimizing under the restriction of the constraint \gls{cvar}, the average tracking error of the digital model achieves 0.0281, which is lower than the tracking error of the baseline (0.0347), as shown in Fig.~\ref{10_cost}. Meanwhile, the \gls{cvar} fluctuates around the constraint bound. The results show that \gls{cppo} converges after 250 training episodes. Meanwhile, constraints and the average tracking error fluctuate slightly throughout training. In particular, \gls{cppo} is stable and effective, since it performs consistently better in all ten training repetitions. It is also worth noting that although the prediction horizon needs to be transmitted to the server, $Z(t)$ is an integer ranging from $1$ to $500$. Thus, the communication overhead for updating $Z(t)$ is negligible compared to the update of the joint angle with high precision. 

\subsection{Ablation Study of C-PPO Algorithm}
\begin{figure}%[p]

  \subfigure[\ \ \ Average packet rate in \\ each \ episode.]{%caption of the subfigure
  \label{domain_reward}%%label for first subfigure
  \includegraphics[scale=0.189]{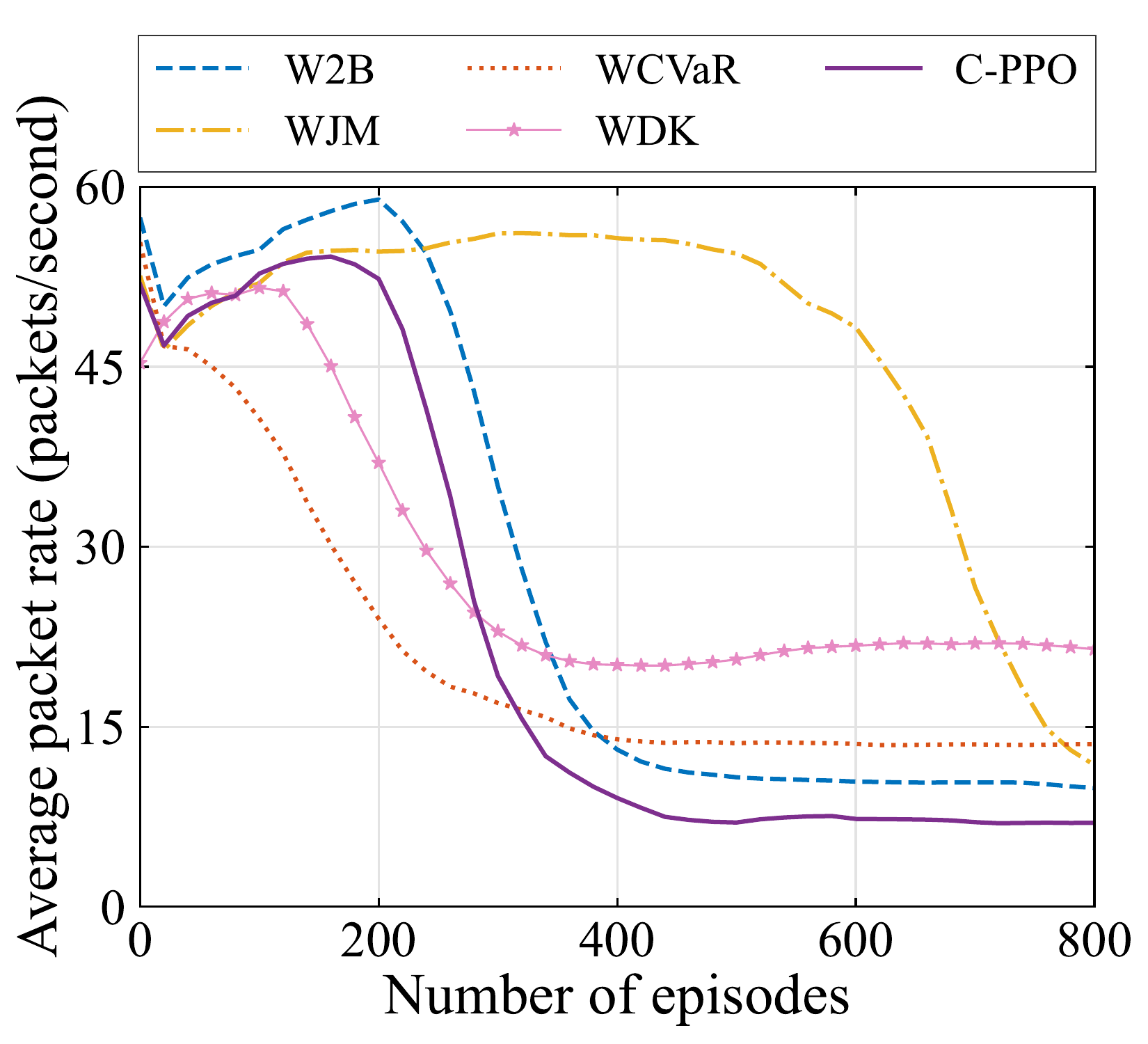}
  }\subfigure[\ \ Average modeling error \\ \ \ \ \ in each episode.]{%caption of the subfigure
  \label{domain_cost}%%label for second subfigure
  \includegraphics[scale=0.189]{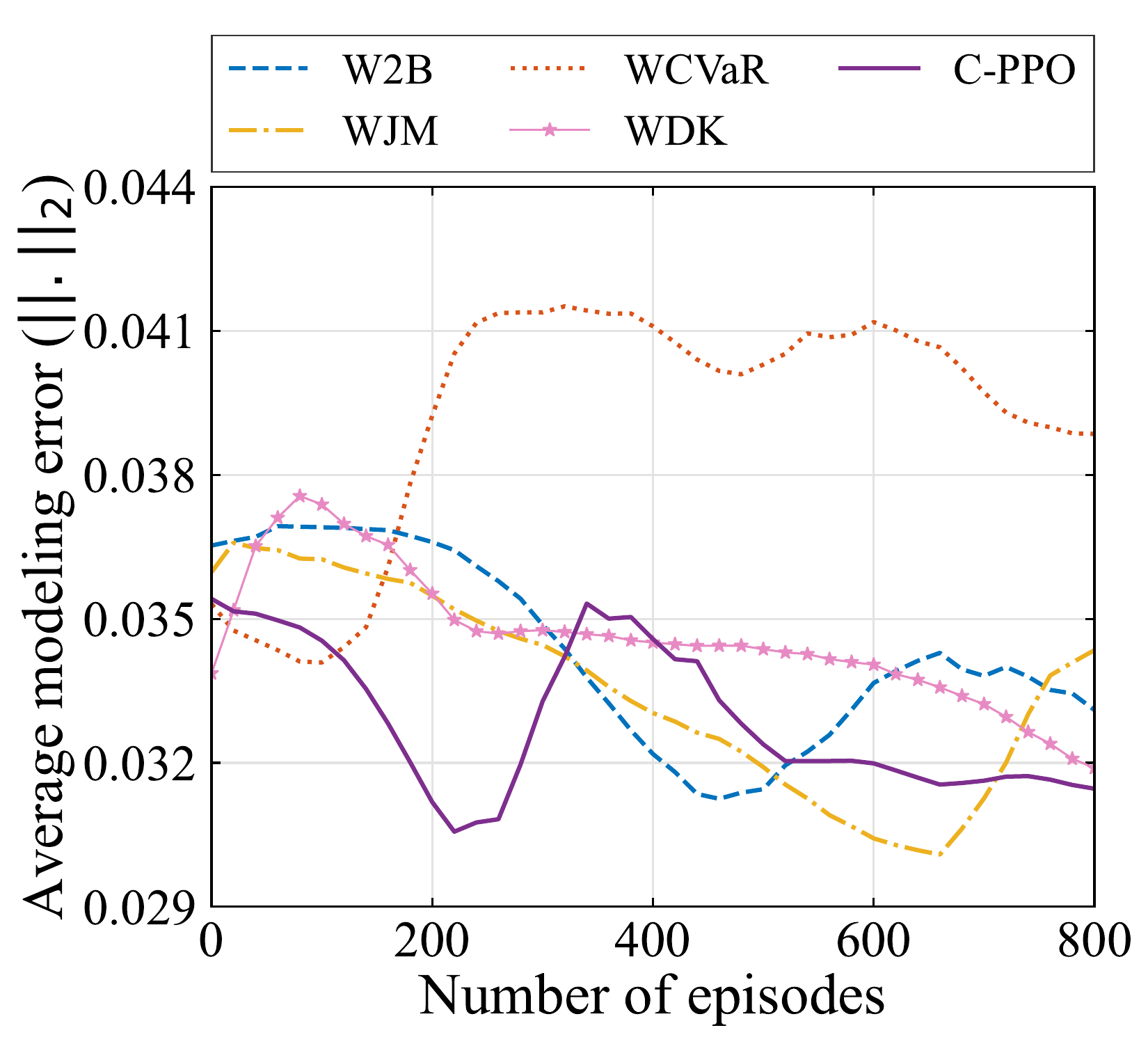}
  }\subfigure[\ \ Average value of \gls{cvar} \\ \ \ \ \ in each episode.]{%caption of the subfigure
  \label{domain_cvar}%%label for second subfigure
  \includegraphics[scale=0.186]{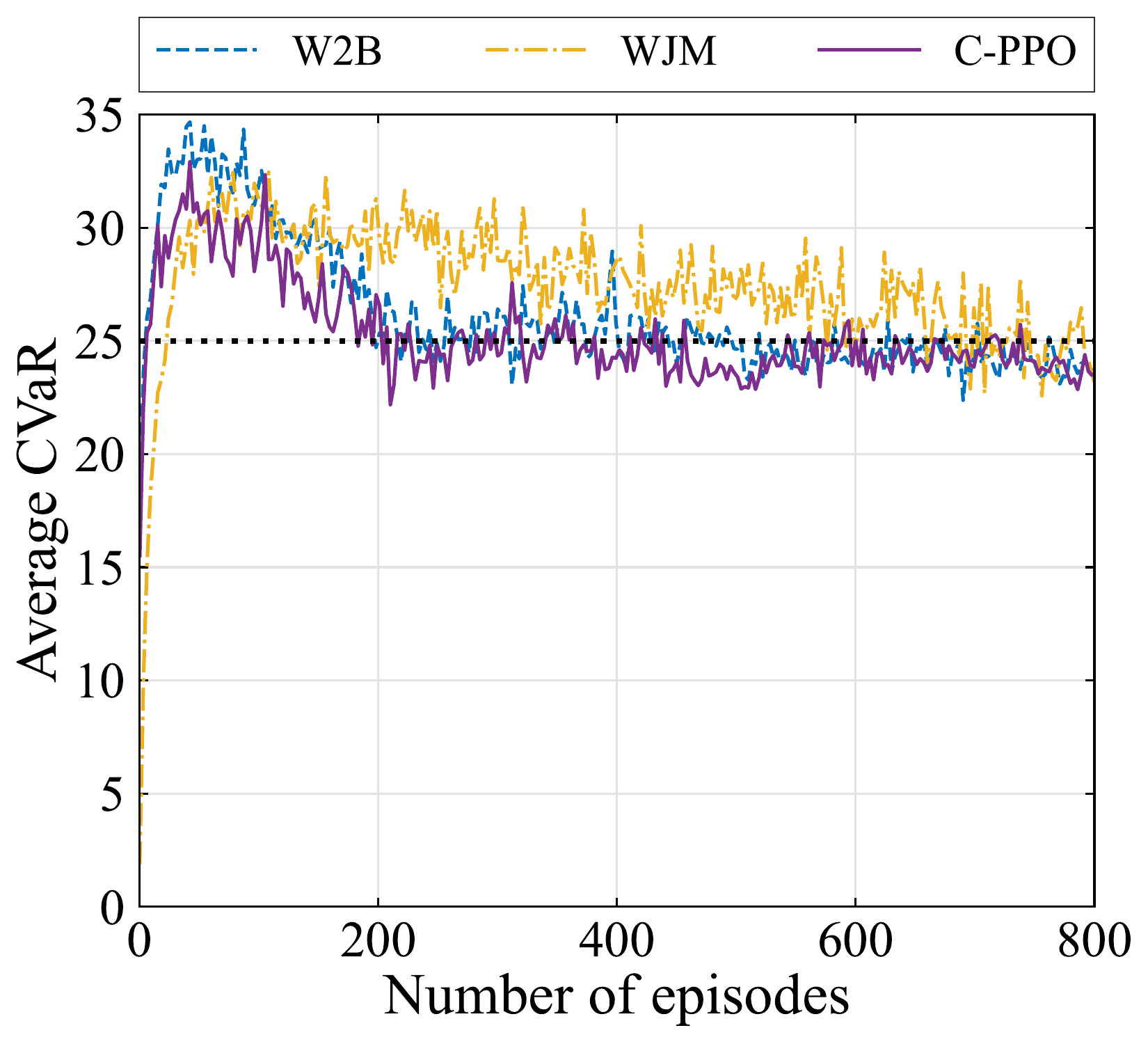}
  }  

  \caption{Ablation Study.} % in the training stage}
  \label{fig: ablation}%% label for entire figure
\end{figure}

The performance of the \gls{cppo} and the four benchmarks are illustrated in Fig.~\ref{fig: ablation}. In general, our \gls{cppo} achieves the best performance in terms of convergence time, average packet rate, and average modeling error. The results in Fig.~\ref{domain_reward} show that the \gls{cppo} can reduce the required average packet rate by around $50$\% compared to the benchmark without domain knowledge, WDK (from $17$ packets/second to $8$ packets/second). By comparing \gls{cppo} with WJM, we can see that the Jacobian matrix can reduce the convergence time by $50$\% (from $800$ episodes to $400$ episodes). In Fig.~\ref{domain_cost}, we evaluate the average modeling errors achieved by different algorithms. From \gls{cppo} and WCVaR, we can observe that by using CVaR as the metric of the modeling error, the average modeling error is reduced from $0.041$ to $0.032$ (around $20$\% reduction), where the modeling error is defined in \eqref{weight_norm}. The results \ref{domain_cvar} show that \gls{cppo}, W2B, and WJM can guarantee the constraint on CVaR of the modeling error, that is, the dashed horizontal line. The other two benchmarks do not consider CVaR, and hence are not shown in this figure. From the performance of W2B in all these figures, we can see that the two-branch neural network converges faster than the fully-connected neural network and achieves better performance in terms of the average packet rate and average modeling error.

\subsection{Validation of Cross-System Design Framework}

\subsubsection{Dynamic scheduling in C-PPO}
\begin{figure}
             \subfigure[Trajectories and instantaneous packet rates]{
              \label{Action_s} %%label for first subfigure
              \includegraphics[scale=0.36]{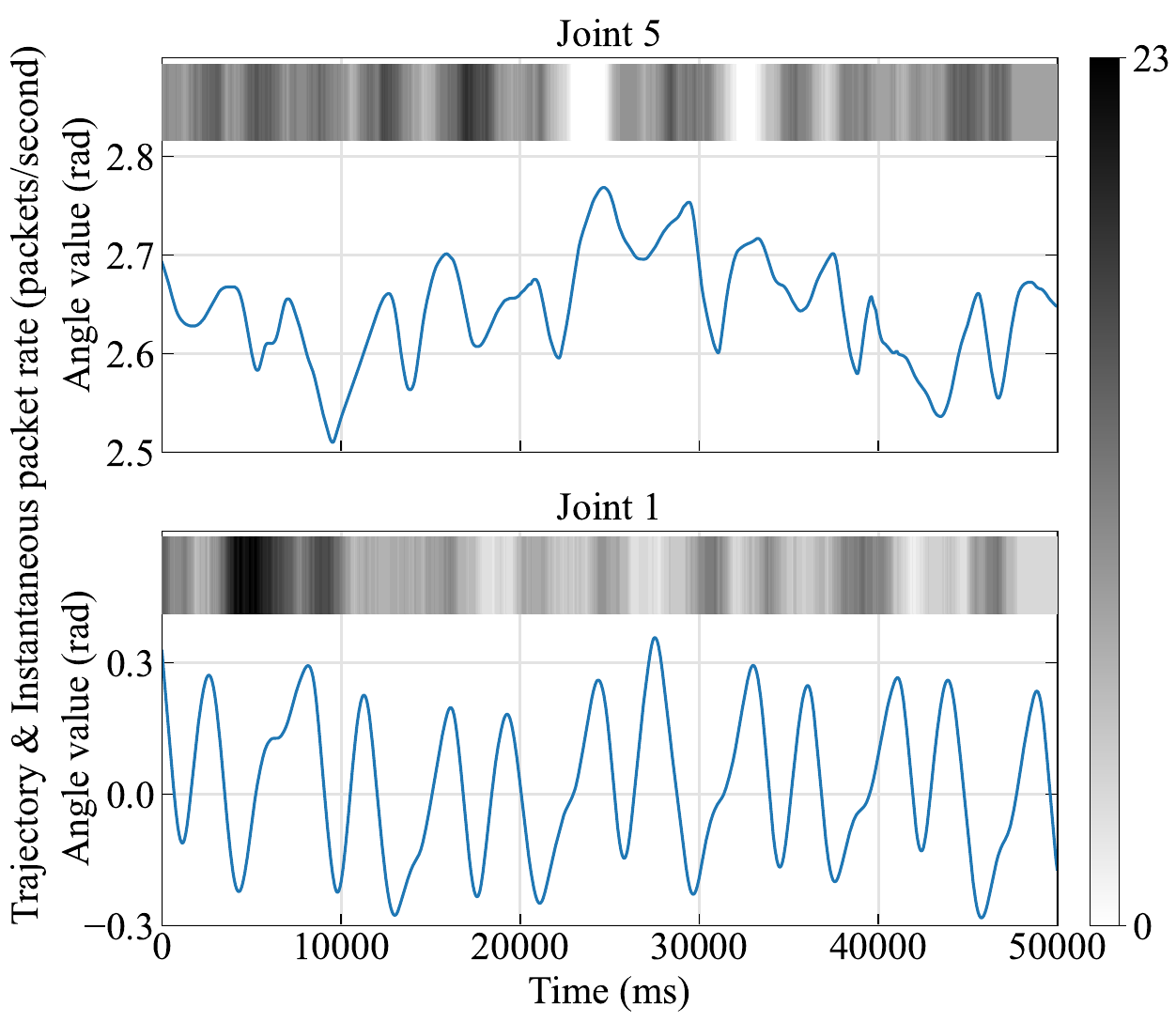}
              }  
              \subfigure[Trajectories and instantaneous prediction horizon]{
              \label{Action_pre} %%label for first subfigure
              \includegraphics[scale=0.289]{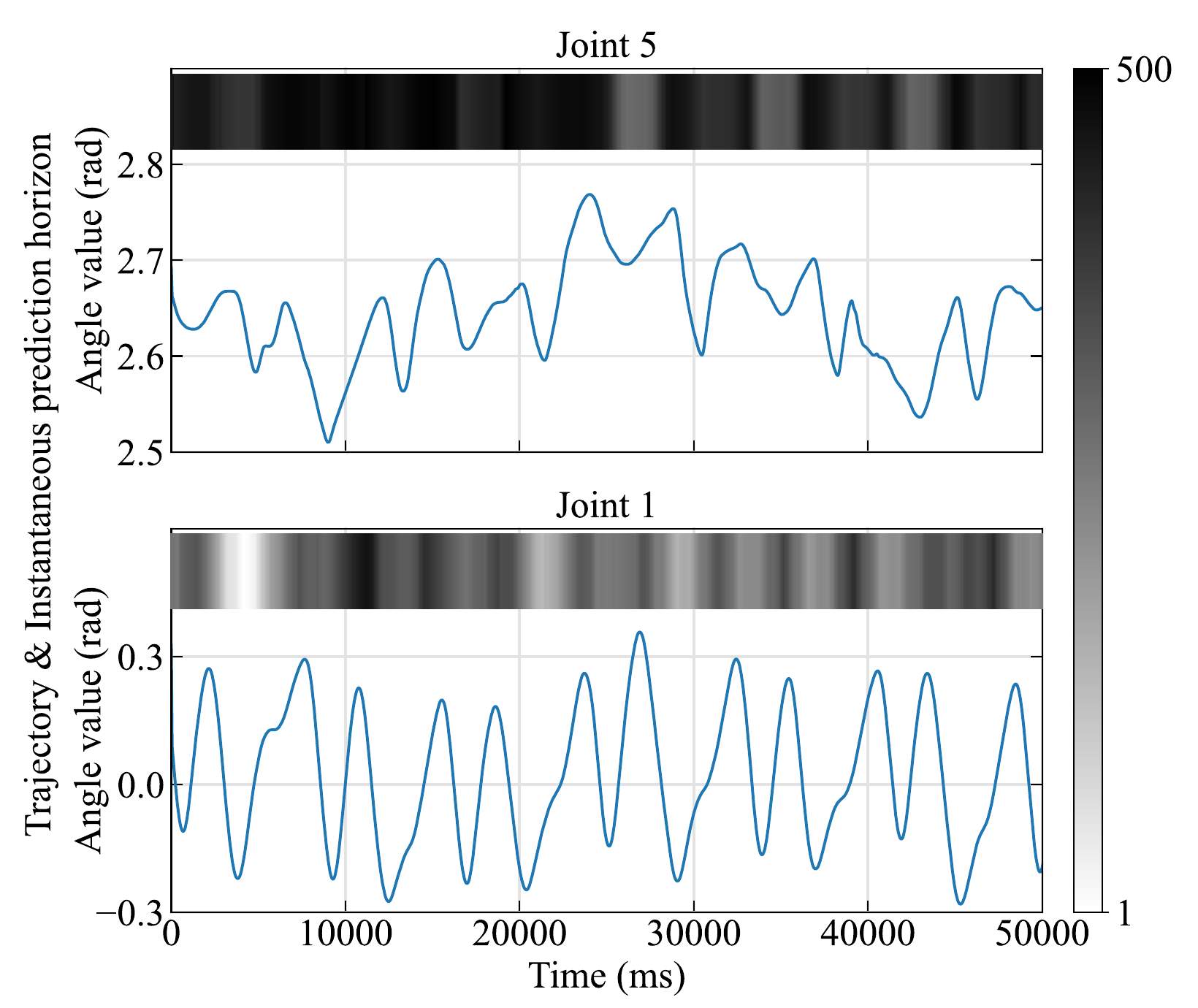}
              }
              
            \caption{Trajectories, instantaneous packet rates and instantaneous prediction horizon of two joints, where joint 5 is close to the end effector, and joint 1 is close to the base of the robotic arm.}
            \label{Action}
\end{figure}

In Fig.~\ref{Action_s}, we provide an example to show how the proposed \gls{cppo} algorithm changes the packet rates according to joint angles. The packet rates are represented by the grayscale intensity. As the grayscale intensity increases from white to black, the packet rates increase from $0$ to $23$ packets/second. The results imply that packet rates are correlated with fluctuations in joint angles. Besides, the joint that is far away from the end effector has higher average packet rates than the joint that is close to the end effector. This is because the modeling error of the end effector is less sensitive to modeling errors of the joints that are closer to it. 

\subsubsection{Dynamic prediction horizon in C-PPO}
The effect of $Z(t)$ has already been demonstrated in the existing literature~\cite{8902186,9681620}. The latency in the communication systems is stochastic, so we need to adjust the prediction horizon to  compensate for the communication latency. In this way, we can reduce the modeling error in the Metaverse. In addition, we also provide the diagram of $Z(t)$ along with the trajectories of the real-world robotic arm. As shown in~Fig.~\ref{Action_pre}, the value of the prediction horizon is depicted through the grayscale intensity, where darker intensities correspond to longer prediction horizons. The results show that the prediction horizon is adjusted according to the mobility of the joints. Moreover, the prediction horizon of joint 5 is much longer than that of joint 1. This is because the modeling accuracy of the end effector is less sensitive to the prediction errors of the joint that is closer to it  compared to the joint that is farther from it.

\subsubsection{CVaR}
\begin{figure}
            \centering
           \includegraphics[scale=0.32]{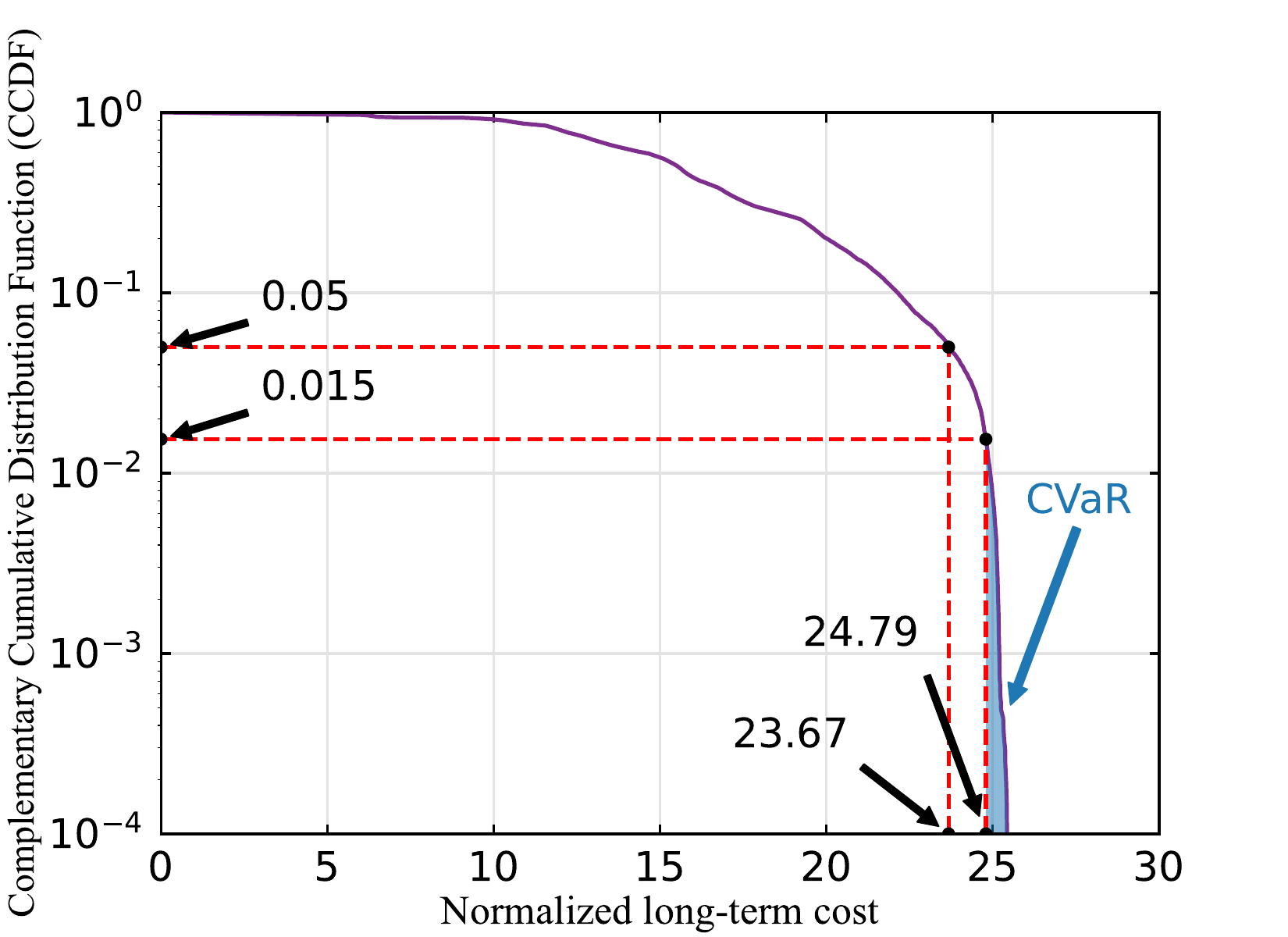}
           \caption{CCDF of the long-term cost.}
           \label{fig: CVaR}
\end{figure}

The \gls{ccdf} of the long-term cost is presented in Fig.~\ref{fig: CVaR}. The results show that with the probability of $98.5$\%, the long-term cost is below the required threshold, which is set to $25$ in the experiment. In addition, the probability ($98.5$\%) is higher than the confidence level ($95$\%).
%In addition, the long-term cost is lower than 23.67 at the original threshold $1-\alpha_{c}=95\%$.
The result also indicates that the proposed \gls{cppo} can significantly reduce the tail probability (i.e., the probability that the long-term cost is higher than the required threshold) of the long-term cost.

\begin{figure}
            \centering
           \includegraphics[scale=0.35]{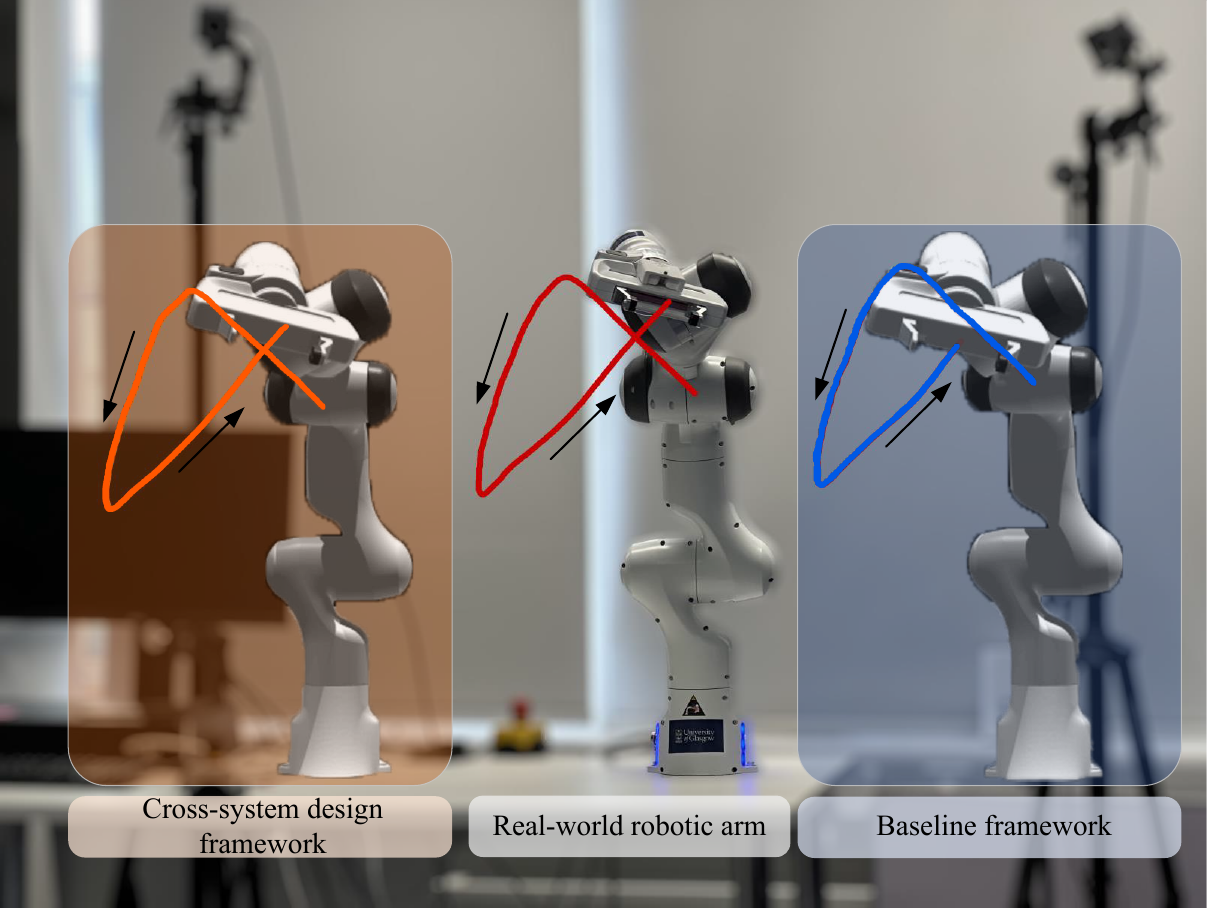}
           \caption{Comparison among the trajectories of the real-world robotic arm (at the centre) and its digital models (on two sides, the left one is designed by the proposed  framework while the right one is implemented by the baseline framework without cross-system design). }
           \label{fig: end_effector}
\end{figure}

\begin{figure}
            \centering
           \includegraphics[scale=0.32]{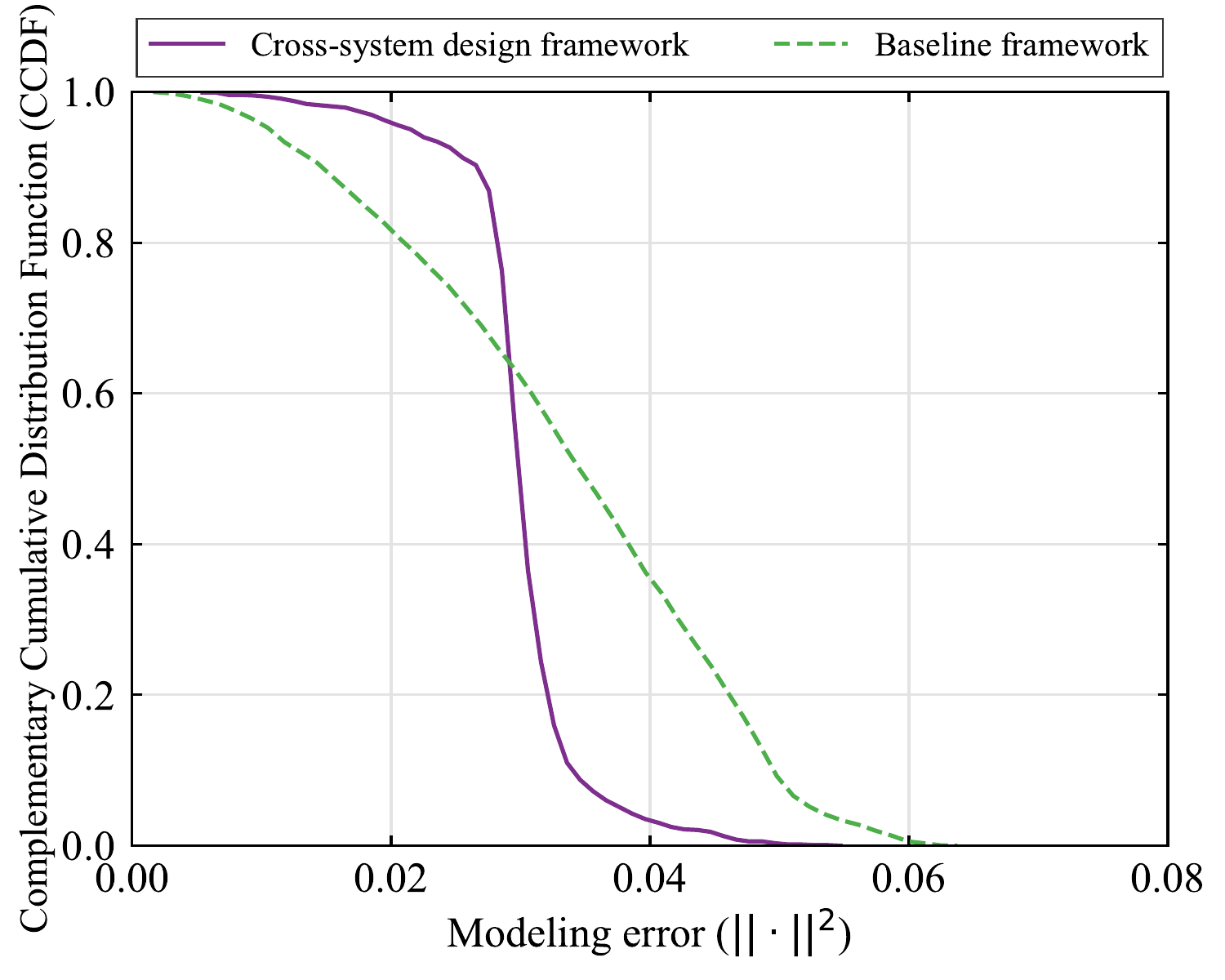}
           \caption{Modeling errors of the proposed cross-system design framework and the baseline framework. }
           \label{fig: cdf}
\end{figure}

\subsubsection{Performance Comparison} We compare our proposed cross-system design framework with a baseline framework: there is no prediction, and all joints transmit data packets in every time slot. In Fig.~\ref{fig: end_effector}, we show the trajectories of the real-world robotic and two digital models obtained from our cross-system design framework and the baseline framework. The results show that with prediction, the cross-system design framework can model the virtual-world robotic arm in a timely manner. Without prediction, the user can recognize the modeling error caused by communication latency. In Fig.~\ref{fig: cdf}, we further illustrate the \gls{ccdf} of modeling errors in the two frameworks. The results demonstrate that the cross-system design framework outperforms the baseline framework in terms of the tail distribution of the modeling error. Besides, with the baseline framework, all the joints transmit packets in every time slot. The packet rate is $5000$~packets/s, which is much higher than the cross-system design framework.

\section{Conclusions}
\label{sec:conclusions}
In this work, we established a task-oriented cross-system design framework to minimize the required packet rate to meet a constraint on the modeling error of a robotic arm in the Metaverse. To optimize the scheduling policy and the prediction horizons, we developed a \gls{cppo} algorithm by integrating domain knowledge into the \gls{ppo}. A prototype was built to evaluate the performance of the \gls{cppo} and the cross-system design framework. Experimental results showed that the domain knowledge helps reduce the required packet rate and the convergence time by up to $50$\%, and the cross-design framework outperforms a baseline framework in terms of the required packet rate and the tail distribution of the modeling error.

\begin{appendices}
\section{Calculation of Jacobian Matrix}\label{app:A}

For notational simplicity, the notations used in the appendix are different from the notations used in the main text. 

\begin{figure}
    \centering
    \includegraphics[scale=0.5]{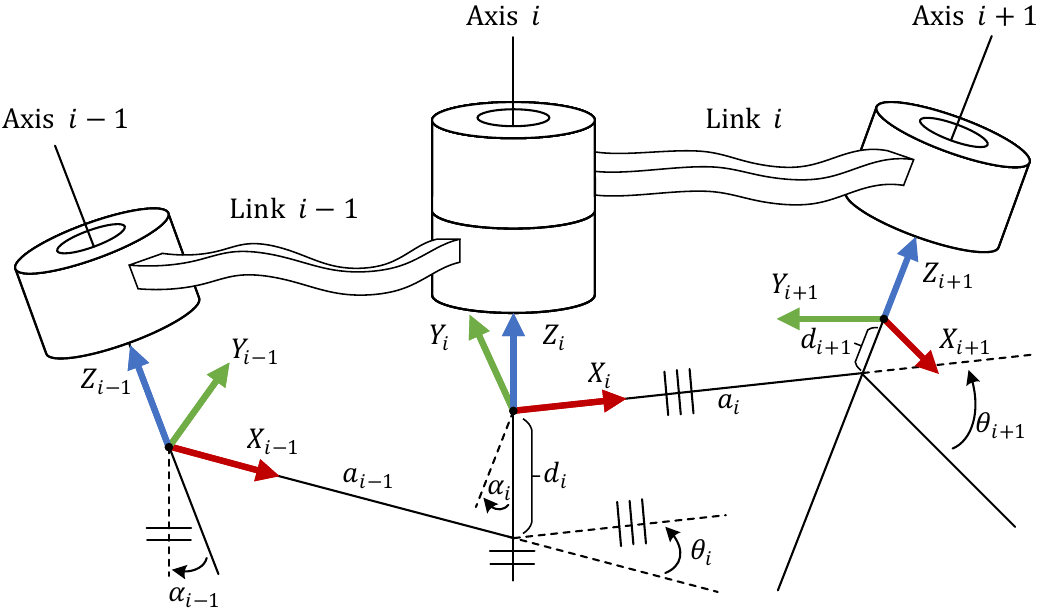}
    \caption{The coordinate frame $\{i\}$ and \gls{dh} parameters.}
    \label{fig: link}
\end{figure}

To obtain the Jacobian matrix defined in Section~\ref{Subsec:CMDP}, one approach is to compute the partial derivatives with respect to each joint angle. The computation complexity of this approach could be high, and we introduce a low-complexity numerical method in this appendix to obtain the Jacobian matrix~\cite{craighead2008using}. For a robotic arm with $I$ rotation joints, the Jacobian matrix can be obtained from the following expression,
\begin{align}\label{eq:jacobian Matrix}
 \mathcal{J} &= \begin{bmatrix} { }_{0}^{0} R\begin{bmatrix} 0\\0\\1\end{bmatrix}\times ({ }_{I}^{0}\xi-{ }_{0}^{0}\xi)& { }_{1}^{0} R\begin{bmatrix} 0\\0\\1\end{bmatrix}\times ({ }_{I}^{0}\xi-{ }_{1}^{0}\xi) & \dots &{ }_{I-1}^{0}R\begin{bmatrix}0\\0\\1\end{bmatrix}\times ({}^{0}_{I}\xi-{}^{0}_{I-1}\xi)\\ 
 { }_{0}^{0}R\begin{bmatrix}0\\0\\1\end{bmatrix}& { }_{1}^{0}R\begin{bmatrix}0\\0\\1\end{bmatrix} & \dots & { }_{I-1}^{0}R\begin{bmatrix}0\\0\\1\end{bmatrix}\end{bmatrix} \in \mathbb{R}^{6 \times I},
\end{align}
where $\times$ is the cross product operation defined by: ${\bf{a}}=[x_1, y_1, z_1]$, ${\bf{b}}=[x_2, y_2, z_2]$, ${\bf{a}} \times {\bf{b}} = [y_1z_2-y_2z_1,\ x_2z_1-x_1z_2,\ x_1y_2-x_2y_1] {}^{\mathrm{T}}$, ${ }_{I-1}^{0}R \in \mathbb{R}^{3 \times 3}$ is the rotation matrix that describes the rotation of the coordinate frame $\{I-1\}$ in the coordinate frame \{0\} which is the base coordinate and ${}^{0}_{I-1}\xi \in \mathbb{R}^{3 \times 1}$ is the translation vector that describes the translation of the origin of the coordinate frame $\{I-1\}$ in the coordinate frame $\{0\}$. In robotics, a coordinate frame is a system of reference used to describe the position and motion of robots in space. As shown in Fig.~\ref{fig: link}, the rotation center of a joint is commonly used as the reference point for setting up the coordinate frame $\{i\}$. Then, by concatenating the rotation matrix and the translation vector, the transformation matrix of ${ }_{I-1}^{0}T$ is expressed as
\begin{align}\label{eq: TM}
{ }_{I-1}^{0}T =
\left[\begin{array}{c|c}
{ }_{I-1}^{0} R & { }_{I-1}^{0}\xi \\
\hline [0, 0, 0] & 1
\end{array}\right] \in \mathbb{R}^{4 \times 4},
\end{align}
which describes the relative position and orientation of coordinate frame $\{I-1\}$ with respect to the coordinate frame $\{ 0\}$.  

One way to obtain the transformation matrix is deriving the ~\gls{dh} parameters~\cite{craighead2008using}. ~\gls{dh} parameters provide a systematic way to describe the position and orientation of each link and joint in the robot in the joint space which is widely used by industrial manufacturers. The relationship between ~\gls{dh} parameters and transformation matrix ${}_{i}^{i-1}T$ can be expressed by

\begin{equation}\label{eq:transformation matrix}
{}_{i}^{i-1}T=\left[\begin{array}{cccc}
\cos \theta_{i} & -\sin \theta_{i} & 0 & a_{i-1} \\
\sin \theta_{i} \cos \alpha_{i-1} & \cos \theta_{i} \cos \alpha_{i-1} & -\sin \alpha_{i-1} & -\sin (\alpha_{i-1}) d_{i} \\
\sin \theta_{i} \sin \alpha_{i-1} & \cos \theta_{i} \sin \alpha_{i-1} & \cos \alpha_{i-1} & \cos (\alpha_{i-1}) d_{i} \\
0 & 0 & 0 & 1
\end{array}\right].
\end{equation}
As shown in Fig~\ref{fig: link}, $\theta_{i}$ is the angle value of $i$-th joint, $\alpha_{i-1}$, $a_{i-1}$, $d_i$ are the constant parameters determined by the mechanical system. Then, the transformation matrix ${}_{i}^{0}T$ between coordinate frames $\{0\}$ and $\{i\}$ can be obtained by the forward kinematic chain~\cite{craighead2008using},
\begin{equation}\label{eq:fk chain}
{}_{i}^{0}T = {}_{1}^{0}T{}_{2}^{1}T{}_{3}^{2}T\cdot\cdot\cdot{}_{i}^{i-1}T
\end{equation}

In our prototype, Franka Emika Panda robotic arm is used. The corresponding ~\gls{dh} parameters are shown in Table~\ref{DH_panda}~\cite{franka_spec}. Thus, by substituting the ~\gls{dh} parameters into (\ref{eq:fk chain}), we can obtain ${}_{i}^{0}T$, ${}_{i}^{0}R$ and ${}_{i}^{0}\xi$, $i= 1, 2,...,I$. Then, by substituting ${}_{i}^{0}R$ and ${}_{i}^{0}\xi$, $i= 1, 2,...,I$, into (\ref{eq:jacobian Matrix}), we can obtain the Jacobian matrix $\mathcal{J}$. This completes the calculation of the Jacobian matrix.

\section{Foundation of Quaternion}
\label{apd: quaternion}
Specifically, imaginary basis vectors follow the multiplication rules as
\begin{align}\label{eq: quat_multi_1}
    \bm{u}_\text{i}^2=\bm{u}_\text{j}^2=\bm{u}_\text{k}^2=-1, \qquad \bm{u}_\text{i}\bm{u}_\text{j}\bm{u}_\text{k}=-1
\end{align}
\begin{align}
    \bm{u}_\text{i}\bm{u}_\text{j}=-\bm{u}_\text{j}\bm{u}_\text{i}=\bm{u}_\text{k}, \qquad \bm{u}_\text{j}\bm{u}_\text{k}=-\bm{u}_\text{k}\bm{u}_\text{j}=\bm{u}_\text{i}, \qquad \bm{u}_\text{k}\bm{u}_\text{i}=-\bm{u}_\text{i}\bm{u}_\text{k}=\bm{u}_\text{j}.
\end{align}
As shown in Fig. 3, according to the unit vector of the rotation axis $\bm{\eta}_\text{c}$ and the rotation angle $\psi_\eta$, the quaternion can be obtained by
\begin{align}
    \bm{q} &= F_{q}(a_\mathrm{c}, b_\mathrm{c}, c_\mathrm{c}, \psi_\eta) \notag \\
    &= \sin(\frac{\psi_\eta}{2})\cdot a_\mathrm{c} \cdot \bm{u}_\text{i}
    + \sin(\frac{\psi_\eta}{2})\cdot b_\mathrm{c} \cdot \bm{u}_\text{j}
    + \sin(\frac{\psi_\eta}{2})\cdot c_\mathrm{c} \cdot \bm{u}_\text{k}
    + \cos(\frac{\psi_\eta}{2}).
\end{align}

\end{appendices}

\begin{table}[H]
\centering
\caption{\gls{dh} PARAMETERS OF FRANKA EMIKA PANDA ROBOTIC ARM}
\label{DH_panda}
\begin{tabular}{|c|c|c|c|c|}
\hline
Joint   & $a$ (m)    & $d$ (m)  & $\alpha$ (rad) & $\theta$ (rad)   \\ \hline
Joint 1 & 0       & 0.333 & 0      &      $\theta_1$     \\ \hline
Joint 2 & 0       & 0     & $-\frac{\pi}{2}$  &      $\theta_2$     \\ \hline
Joint 3 & 0       & 0.316 & $\frac{\pi}{2}$   &      $\theta_3$      \\ \hline
Joint 4 & 0.0825  & 0     & $\frac{\pi}{2}$  &      $\theta_4$      \\ \hline
Joint 5 & -0.0825 & 0.384 & $-\frac{\pi}{2}$  &      $\theta_5$      \\ \hline
Joint 6 & 0       & 0     & $\frac{\pi}{2}$   &      $\theta_6$      \\ \hline
Joint 7 & 0.088   & 0     & $\frac{\pi}{2}$   &      $\theta_7$      \\ \hline
\end{tabular}
\end{table}

\normalem
\bibliography{IEEEabrv,main}

% Generated by IEEEtran.bst, version: 1.12 (2007/01/11)
\begin{thebibliography}{10}
\providecommand{\url}[1]{#1}
\csname url@samestyle\endcsname
\providecommand{\newblock}{\relax}
\providecommand{\bibinfo}[2]{#2}
\providecommand{\BIBentrySTDinterwordspacing}{\spaceskip=0pt\relax}
\providecommand{\BIBentryALTinterwordstretchfactor}{4}
\providecommand{\BIBentryALTinterwordspacing}{\spaceskip=\fontdimen2\font plus
\BIBentryALTinterwordstretchfactor\fontdimen3\font minus \fontdimen4\font\relax}
\providecommand{\BIBforeignlanguage}[2]{{%
\expandafter\ifx\csname l@#1\endcsname\relax
\typeout{** WARNING: IEEEtran.bst: No hyphenation pattern has been}%
\typeout{** loaded for the language `#1'. Using the pattern for}%
\typeout{** the default language instead.}%
\else
\language=\csname l@#1\endcsname
\fi
#2}}
\providecommand{\BIBdecl}{\relax}
\BIBdecl

\bibitem{lee2021all}
\BIBentryALTinterwordspacing
L.-H. Lee, T.~Braud, P.~Zhou, L.~Wang, D.~Xu, Z.~Lin, and et~al, ``All one needs to know about {M}etaverse: A complete survey on technological singularity, virtual ecosystem, and research agenda,'' 2021. [Online]. Available: \url{https://arxiv.org/abs/2110.05352}
\BIBentrySTDinterwordspacing

\bibitem{583063}
S.~Ellis, F.~Breant, B.~Manges, R.~Jacoby, and B.~Adelstein, ``Factors influencing operator interaction with virtual objects viewed via head-mounted see-through displays: viewing conditions and rendering latency,'' in \emph{Proc. IEEE Annu. Int. Symp. Virtual Real. (VRAIS)}, 1997, pp. 138--145.

\bibitem{laaki2019prototyping}
H.~Laaki, Y.~Miche, and K.~Tammi, ``Prototyping a digital twin for real time remote control over mobile networks: Application of remote surgery,'' \emph{IEEE Access}, vol.~7, pp. 20\,325--20\,336, 2019.

\bibitem{3GPP}
``\emph{Study on scenarios and requirements for next generation access technologies},'' \emph{\rm{document 3GPP, TSG RAN TR38.913 R14}}, Jun. 2017.

\bibitem{girgis2021predictive}
A.~M. Girgis, J.~Park, M.~Bennis, and M.~Debbah, ``Predictive control and communication co-design via two-way gaussian process regression and {AoI}-aware scheduling,'' \emph{IEEE Trans. Commun.}, vol.~69, no.~10, pp. 7077--7093, 2021.

\bibitem{hou2019prediction}
Z.~Hou, C.~She, Y.~Li, L.~Zhuo, and B.~Vucetic, ``Prediction and communication co-design for ultra-reliable and low-latency communications,'' \emph{IEEE Trans. Wireless Commun.}, vol.~19, no.~2, pp. 1196--1209, 2019.

\bibitem{15450}
\BIBentryALTinterwordspacing
E.~Fountoulakis, M.~Codreanu, A.~Ephremides, and N.~Pappas, ``Joint sampling and transmission policies for minimizing cost under {AoI} constraints,'' 2021. [Online]. Available: \url{https://arxiv.org/abs/2103.15450}
\BIBentrySTDinterwordspacing

\bibitem{8812616}
Y.~Sun, Y.~Polyanskiy, and E.~Uysal, ``Sampling of the wiener process for remote estimation over a channel with random delay,'' \emph{IEEE Trans. Inf. Theory}, vol.~66, no.~2, pp. 1118--1135, 2020.

\bibitem{schulman2017proximal}
\BIBentryALTinterwordspacing
J.~Schulman, F.~Wolski, P.~Dhariwal, A.~Radford, and O.~Klimov, ``Proximal policy optimization algorithms,'' 2017. [Online]. Available: \url{https://arxiv.org/abs/1707.06347}
\BIBentrySTDinterwordspacing

\bibitem{liang2018accelerated}
\BIBentryALTinterwordspacing
Q.~Liang, F.~Que, and E.~Modiano, ``Accelerated primal-dual policy optimization for safe reinforcement learning,'' 2018. [Online]. Available: \url{https://arxiv.org/abs/1802.06480}
\BIBentrySTDinterwordspacing

\bibitem{01629}
\BIBentryALTinterwordspacing
Y.~Chow, M.~Ghavamzadeh, L.~Janson, and M.~Pavone, ``Risk-constrained reinforcement learning with percentile risk criteria,'' 2015. [Online]. Available: \url{https://arxiv.org/abs/1512.01629}
\BIBentrySTDinterwordspacing

\bibitem{xu2021crpo}
T.~Xu, Y.~Liang, and G.~Lan, ``{CRPO}: A new approach for safe reinforcement learning with convergence guarantee,'' in \emph{Proc. Int. Conf. on Mach. Learn. (ICML)}, 2021, pp. 11\,480--11\,491.

\bibitem{gu2021knowledge}
Z.~Gu, C.~She, W.~Hardjawana, S.~Lumb, D.~McKechnie, T.~Essery, and et~al, ``{Knowledge-assisted deep reinforcement learning in 5G scheduler design: From theoretical framework to implementation},'' \emph{IEEE J. Sel. Areas Commun.}, vol.~39, no.~7, pp. 2014--2028, 2021.

\bibitem{she2021tutorial}
C.~She, C.~Sun, Z.~Gu, Y.~Li, C.~Yang, H.~V. Poor, and et~al, ``{A Tutorial on ultrareliable and low-latency communications in 6G: integrating domain knowledge into deep learning},'' \emph{Proc. IEEE}, vol. 109, no.~3, pp. 204--246, 2021.

\bibitem{stephenson2003snow}
N.~Stephenson, \emph{Snow crash: A novel}.\hskip 1em plus 0.5em minus 0.4em\relax Spectra, 2003.

\bibitem{01672}
\BIBentryALTinterwordspacing
V.-P. Bui, S.~R. Pandey, F.~Chiariotti, and P.~Popovski, ``Game networking and its evolution towards supporting {M}etaverse through the 6{G} wireless systems,'' 2023. [Online]. Available: \url{https://arxiv.org/abs/2302.01672}
\BIBentrySTDinterwordspacing

\bibitem{06838}
\BIBentryALTinterwordspacing
M.~Xu, D.~Niyato, B.~Wright, H.~Zhang, J.~Kang, Z.~Xiong, and et~al, ``{EPViSA}: Efficient auction design for real-time physical-virtual synchronization in the {M}etaverse,'' 2022. [Online]. Available: \url{https://arxiv.org/abs/2211.06838}
\BIBentrySTDinterwordspacing

\bibitem{9865226}
Y.~Han, D.~Niyato, C.~Leung, D.~I. Kim, K.~Zhu, S.~Feng, X.~Shen, and C.~Miao, ``A dynamic hierarchical framework for iot-assisted digital twin synchronization in the metaverse,'' \emph{IEEE Internet Things J.}, vol.~10, no.~1, pp. 268--284, 2023.

\bibitem{01512}
\BIBentryALTinterwordspacing
S.~K. Jagatheesaperumal, K.~Ahmad, A.~Al-Fuqaha, and J.~Qadir, ``Advancing education through extended reality and internet of everything enabled {M}etaverses: Applications, challenges, and open issues,'' 2022. [Online]. Available: \url{https://arxiv.org/abs/2207.01512}
\BIBentrySTDinterwordspacing

\bibitem{9681718}
H.~Tataria, M.~Shafi, M.~Dohler, and S.~Sun, ``Six critical challenges for 6{G} wireless systems: A summary and some solutions,'' \emph{IEEE Vehicular Technol. Mag.}, vol.~17, no.~1, pp. 16--26, 2022.

\bibitem{multi-tier}
\BIBentryALTinterwordspacing
K.~Wang, J.~Jin, Y.~Yang, T.~Zhang, A.~Nallanathan, C.~Tellambura, and B.~Jabbari, ``Task offloading with multi-tier computing resources in next generation wireless networks,'' 2022. [Online]. Available: \url{https://arxiv.org/abs/2205.13866}
\BIBentrySTDinterwordspacing

\bibitem{khan2023metaverse}
\BIBentryALTinterwordspacing
L.~U. Khan, M.~Guizani, D.~Niyato, A.~Al-Fuqaha, and M.~Debbah, ``Metaverse for wireless systems: Architecture, advances, standardization, and open challenges,'' 2023. [Online]. Available: \url{https://arxiv.org/abs/2301.11441}
\BIBentrySTDinterwordspacing

\bibitem{8070468}
C.~She, C.~Yang, and T.~Q.~S. Quek, ``Cross-layer optimization for ultra-reliable and low-latency radio access networks,'' \emph{IEEE Trans. Wireless Commun.}, vol.~17, no.~1, pp. 127--141, 2018.

\bibitem{glaessgen2012digital}
E.~Glaessgen and D.~Stargel, ``The digital twin paradigm for future {NASA} and {U.S.} air force vehicles,'' in \emph{Proc. 53rd AIAA/ASME/ASCE/AHS/ASC Struct. Struct. Dyn.}, 2012, pp. 1--14.

\bibitem{13445}
\BIBentryALTinterwordspacing
N.~H. Chu, D.~N. Nguyen, D.~T. Hoang, K.~T. Phan, E.~Dutkiewicz, D.~Niyato, and et~al, ``Dynamic resource allocation for {M}etaverse applications with deep reinforcement learning,'' 2023. [Online]. Available: \url{https://arxiv.org/abs/2302.13445}
\BIBentrySTDinterwordspacing

\bibitem{14686}
\BIBentryALTinterwordspacing
O.~Hashash, C.~Chaccour, W.~Saad, K.~Sakaguchi, and T.~Yu, ``Towards a decentralized {M}etaverse: Synchronized orchestration of digital twins and sub-{M}etaverses,'' 2022. [Online]. Available: \url{https://arxiv.org/abs/2211.14686}
\BIBentrySTDinterwordspacing

\bibitem{03300}
\BIBentryALTinterwordspacing
S.~Zeng, Z.~Li, H.~Yu, Z.~Zhang, L.~Luo, B.~Li, and et~al, ``{HFedMS}: Heterogeneous federated learning with memorable data semantics in industrial {Metaverse},'' 2022. [Online]. Available: \url{https://arxiv.org/abs/2211.03300}
\BIBentrySTDinterwordspacing

\bibitem{9491087}
Y.~Lu, S.~Maharjan, and Y.~Zhang, ``Adaptive edge association for wireless digital twin networks in 6{G},'' \emph{IEEE Internet Things J.}, vol.~8, no.~22, pp. 16\,219--16\,230, 2021.

\bibitem{hashash2022edge}
O.~Hashash, C.~Chaccour, and W.~Saad, ``Edge continual learning for dynamic digital twins over wireless networks,'' 2022.

\bibitem{yu20236g}
J.~Yu, A.~Alhilal, P.~Hui, and D.~H.~K. Tsang, ``6{G} mobile-edge empowered metaverse: Requirements, technologies, challenges and research directions,'' 2023.

\bibitem{de2023goal}
P.~M.~d. Santana, N.~Marchenko, B.~Soret, and P.~Popovski, ``Goal-oriented wireless communication for a remotely controlled autonomous guided vehicle,'' \emph{IEEE Wireless Commun. Lett.}, 2023.

\bibitem{shao2021learning}
J.~Shao, Y.~Mao, and J.~Zhang, ``Learning task-oriented communication for edge inference: An information bottleneck approach,'' \emph{IEEE J. Sel. Areas Commun.}, vol.~40, no.~1, pp. 197--211, 2021.

\bibitem{00969}
\BIBentryALTinterwordspacing
D.~Wen, P.~Liu, G.~Zhu, Y.~Shi, J.~Xu, Y.~C. Eldar, and et~al, ``Task-oriented sensing, computation, and communication integration for multi-device edge {AI},'' 2022. [Online]. Available: \url{https://arxiv.org/abs/2207.00969}
\BIBentrySTDinterwordspacing

\bibitem{01471}
\BIBentryALTinterwordspacing
W.~Yu, T.~J. Chua, and J.~Zhao, ``User-centric heterogeneous-action deep reinforcement learning for virtual reality in the {M}etaverse over wireless networks,'' 2023. [Online]. Available: \url{https://arxiv.org/abs/2302.01471}
\BIBentrySTDinterwordspacing

\bibitem{jagatheesaperumal2023semantic}
\BIBentryALTinterwordspacing
S.~K. Jagatheesaperumal, Z.~Yang, Q.~Yang, C.~Huang, W.~Xu, M.~Shikh-Bahaei, and Z.~Zhang, ``Semantic-aware digital twin for metaverse: A comprehensive review,'' 2023. [Online]. Available: \url{https://arxiv.org/abs/2305.18304}
\BIBentrySTDinterwordspacing

\bibitem{scaglia2010linear}
G.~Scaglia, A.~Rosales, L.~Quintero, V.~Mut, and R.~Agarwal, ``A linear-interpolation-based controller design for trajectory tracking of mobile robots,'' \emph{Control Eng. Pract.}, vol.~18, no.~3, pp. 318--329, 2010.

\bibitem{zhou2021informer}
H.~Zhou, S.~Zhang, J.~Peng, S.~Zhang, J.~Li, H.~Xiong, and et~al, ``Informer: Beyond efficient transformer for long sequence time-series forecasting,'' in \emph{Proc. AAAI Conf. Artif. Intell. (AAAI)}, vol.~35, no.~12, 2021, pp. 11\,106--11\,115.

\bibitem{ross1995stochastic}
S.~M. Ross, \emph{Stochastic processes}.\hskip 1em plus 0.5em minus 0.4em\relax John Wiley \& Sons, 1995.

\bibitem{craig2006introduction}
J.~J. Craig, \emph{Introduction to robotics}.\hskip 1em plus 0.5em minus 0.4em\relax Pearson Education, 2006.

\bibitem{motion}
``{Motion Generation},'' \url{https://docs.omniverse.nvidia.com/app_isaacsim/app_isaacsim/ext_omni_isaac_motion_generation.html#isaac-sim-motion-generation}, (accessed Mar. 2023).

\bibitem{05849}
\BIBentryALTinterwordspacing
A.~Tewari, J.~Thies, B.~Mildenhall, P.~Srinivasan, E.~Tretschk, W.~Yifan, and et~al, ``Advances in neural rendering,'' 2021. [Online]. Available: \url{https://arxiv.org/abs/2111.05849}
\BIBentrySTDinterwordspacing

\bibitem{kuipers1999quaternions}
J.~B. Kuipers, \emph{Quaternions and rotation sequences: a primer with applications to orbits, aerospace, and virtual reality}.\hskip 1em plus 0.5em minus 0.4em\relax Princeton Univ. Press, 1999.

\bibitem{Diebel_2006_representing}
J.~Diebel, ``Representing attitude: Euler angles, unit quaternions, and rotation vectors,'' \emph{Matrix}, vol.~58, 01 2006.

\bibitem{2017Proximal}
J.~Schulman, F.~Wolski, P.~Dhariwal, A.~Radford, and O.~Klimov, ``Proximal policy optimization algorithms,'' 2017.

\bibitem{2017Emergence}
N.~Heess, T.~B. Dhruva, S.~Sriram, J.~Lemmon, and D.~Silver, ``Emergence of locomotion behaviours in rich environments,'' 2017.

\bibitem{proximal}
``{Proximal Policy Optimization},'' \url{https://spinningup.openai.com/en/latest/algorithms/ppo.html}, (accessed Jan. 2020).

\bibitem{craighead2008using}
J.~Craighead, J.~Burke, and R.~Murphy, ``Using the {U}nity game engine to develop sarge: A case study,'' in \emph{Proc. Simul. Workshop Int. Conf. Intell. Robots Syst. (IROS Workshops)}, vol. 4552, 2008.

\bibitem{multinomial}
``{torch.multinomial - Pytorch 2.0},'' \url{https://pytorch.org/docs/stable/generated/torch.multinomial.html}, (accessed Jul. 2023).

\bibitem{feyzabadi2013human}
S.~Feyzabadi, S.~Straube, M.~Folgheraiter, E.~A. Kirchner, S.~K. Kim, and J.~C. Albiez, ``Human force discrimination during active arm motion for force feedback design,'' \emph{IEEE Trans. Haptics}, vol.~6, no.~3, pp. 309--319, 2013.

\bibitem{rockafellar2000optimization}
R.~T. Rockafellar and S.~Uryasev, ``Optimization of conditional value-at-risk,'' \emph{J. risk}, vol.~2, pp. 21--42, 2000.

\bibitem{stable-baselines3}
\BIBentryALTinterwordspacing
A.~Raffin, A.~Hill, A.~Gleave, A.~Kanervisto, M.~Ernestus, and N.~Dormann, ``Stable-baselines3: Reliable reinforcement learning implementations,'' \emph{J. Mach. Learn. Res.}, vol.~22, no. 268, pp. 1--8, 2021. [Online]. Available: \url{http://jmlr.org/papers/v22/20-1364.html}
\BIBentrySTDinterwordspacing

\bibitem{libfranka}
F.~E. GmbH, ``{{Franka Control Interface Documentation}},'' \url{https://frankaemika.github.io/docs/index.html}, (accessed Sep. 2, 2022).

\bibitem{web_optitrack}
I.~D.~O. NaturalPoint, ``{OptiTrack Prime 13},'' \url{https://www.optitrack.com/cameras/primex-13.html}, (accessed: Jan. 10, 2023).

\bibitem{ang2005pid}
K.~H. Ang, G.~Chong, and Y.~Li, ``{PID} control system analysis, design, and technology,'' \emph{IEEE Trans. Control Syst. Technol.}, vol.~13, no.~4, pp. 559--576, 2005.

\bibitem{gym}
``{Isaac Gym},'' \url{https://developer.nvidia.com/isaac-gym}, (accessed Jul. 2022).

\bibitem{8902186}
Z.~Hou, C.~She, Y.~Li, L.~Zhuo, and B.~Vucetic, ``Prediction and communication co-design for ultra-reliable and low-latency communications,'' \emph{IEEE Trans. Wireless Commun.}, vol.~19, no.~2, pp. 1196--1209, 2020.

\bibitem{9681620}
Z.~Hou, C.~She, Y.~Li, D.~Niyato, M.~Dohler, and B.~Vucetic, ``Intelligent communications for tactile internet in 6{G}: Requirements, technologies, and challenges,'' \emph{IEEE Commun. Mag.}, vol.~59, no.~12, pp. 82--88, 2021.

\bibitem{franka_spec}
F.~E. GmbH, ``{Robot and interface specifications},'' \url{https://frankaemika.github.io/docs/control_parameters.html}, (accessed Mar. 21, 2023).

\end{thebibliography}
\bibliographystyle{IEEEtran}

\end{document}